\documentclass[journal]{IEEEtran}
\usepackage{cite}
\usepackage{url}
\usepackage{multirow}
\usepackage{color}
\usepackage[hidelinks]{hyperref}
\usepackage{breakurl}
\usepackage{amsmath,bm}
\usepackage{array}
\usepackage{amssymb,amsfonts}
\usepackage{booktabs}
\usepackage[all,arc]{xy}
\usepackage{enumerate}
\usepackage{mathrsfs}
\usepackage[pass]{geometry}
\usepackage{graphicx}
\usepackage[caption=false,font=footnotesize]{subfig}
\usepackage{balance}
\usepackage{algorithm}
\usepackage{lipsum}
\usepackage{enumitem}
\usepackage{textcomp}
\usepackage{soul}
\def\BibTeX{{\rm B\kern-.05em{\sc i\kern-.025em b}\kern-.08em
    T\kern-.1667em\lower.7ex\hbox{E}\kern-.125emX}}
\begin{document}
%
%
\title{Deep Learning Algorithms for Bearing Fault Diagnostics -- A Comprehensive Review}
\author{{Shen Zhang}, \IEEEmembership{Student Member, IEEE},
{Shibo Zhang}, \IEEEmembership{Student Member, IEEE}, \\{Bingnan Wang}, \IEEEmembership{Senior Member, IEEE}, and {Thomas G. Habetler}, \IEEEmembership{Fellow, IEEE}
\thanks{Shen Zhang and Bingnan Wang are with Mitsubishi Electric Research Laboratories, 201 Broadway, Cambridge, MA 02139, USA (e-mail: \{szhang, bwang\}@merl.com)}
\thanks{Shen Zhang Thomas G. Habetler are with School of Electrical and Computer Engineering, Georgia Institute of Technology, Atlanta, GA 30332, USA (e-mail: \{shenzhang, tom.habetler\}@gatech.edu)}
\thanks{Shibo Zhang is with Department of Computer Science, Northwestern University, Evanston, IL 60201, USA (e-mail: shibozhang2015@u.northwestern.edu)}
\thanks{The work is supported by Mitsubishi Electric Research Laboratories (MERL). Part of the work by Shen Zhang was done during his internship at MERL.}
\thanks{Corresponding author: Bingnan Wang (e-mail: bwang@merl.com).}}

\maketitle

\begin{abstract}
In this survey paper, we systematically summarize existing literature on bearing fault diagnostics with deep learning (DL) algorithms. While conventional machine learning (ML) methods, including artificial neural network, principal component analysis, support vector machines, etc., 
have been successfully applied to the detection and categorization of bearing faults for decades, recent developments in DL algorithms in the last five years have sparked renewed interest in both industry and academia for intelligent machine health monitoring. In this paper, we first provide a brief review of conventional ML methods, before taking a deep dive into the state-of-the-art DL algorithms for bearing fault applications. Specifically, the superiority of DL based methods are analyzed in terms of fault feature extraction and classification performances; many new functionalities enabled by DL techniques are also summarized. In addition, to obtain a more intuitive insight, a comparative study is conducted on the classification accuracy of different algorithms utilizing the open source Case Western Reserve University (CWRU) bearing dataset. Finally, to facilitate the transition on applying various DL algorithms to bearing fault diagnostics, detailed recommendations and suggestions are provided for specific application conditions. Future research directions to further enhance the performance of DL algorithms on health monitoring are also discussed.
\end{abstract}



\section{Introduction}
\label{sec:introduction}
Electric machines are widely employed in a variety of industry applications and electrified transportation systems. For certain applications these machines may operate under unfavorable conditions, such as high ambient temperature, high moisture and overload, which can eventually result in motor malfunctions that lead to high maintenance costs, severe financial losses, and safety hazards \cite{CT1,MEHB,TGH2}. The malfunction of electric machines can be generally attributed to various faults of different categories, including drive inverter failures, stator winding insulation breakdown, bearing faults and air gap eccentricity. Several surveys regarding the likelihood of induction machine failures conducted by the IEEE Industry Application Society (IEEE-IAS) \cite{IAS1,IAS2,IAS3} and the Japan Electrical Manufacturers' Association (JEMA) \cite{JEMA} reveal that bearing fault is the most common fault type and is responsible for 30\% to 40\% of all the machine failures. 

\begin{figure}[!t]
\centering
\includegraphics[width=3.2in]{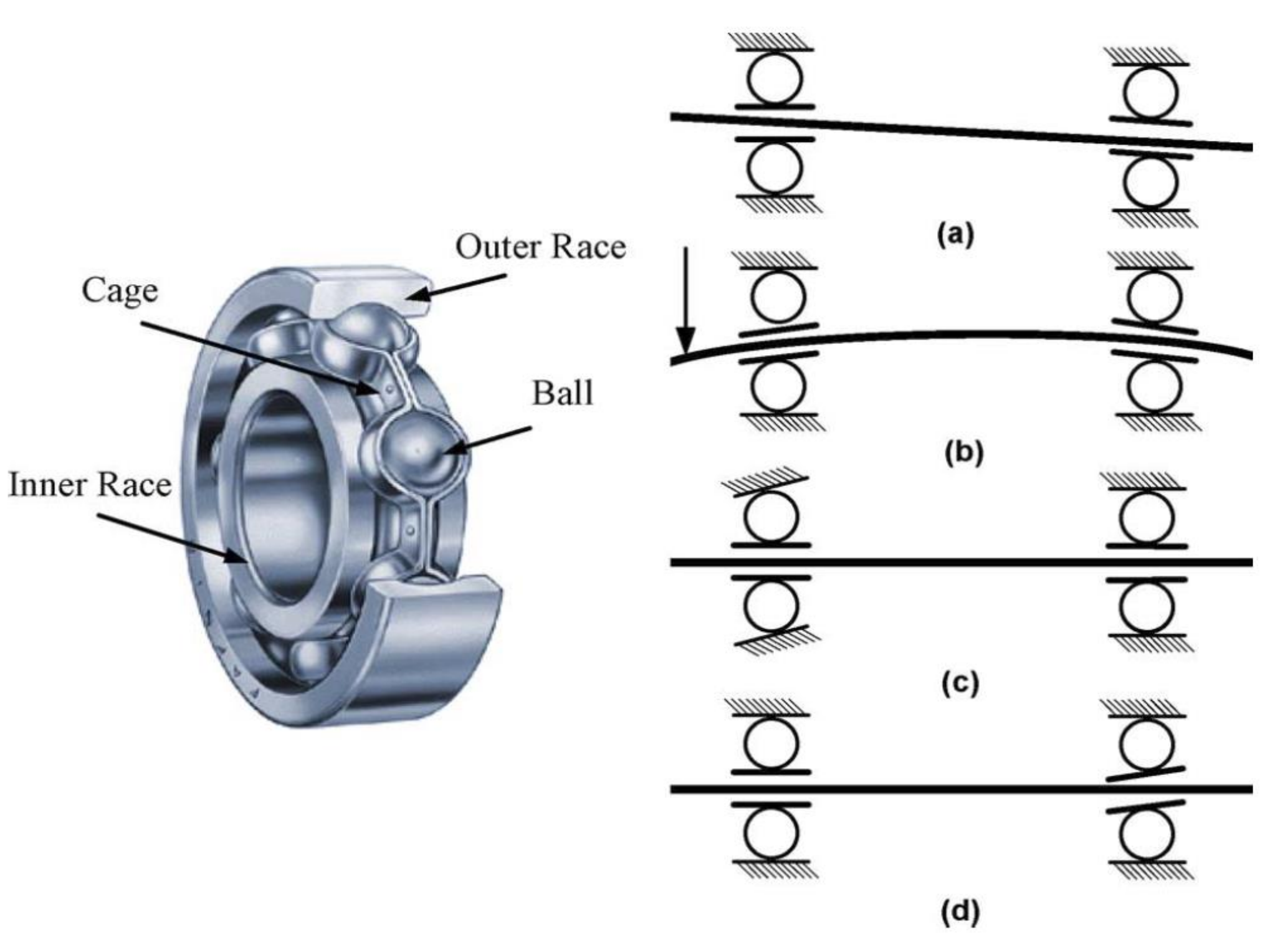}
\caption{Structure of a rolling-element bearing with four types of common scenarios of misalignment that are likely to cause bearing failures: (a) misalignment (out-of-line), (b) shaft deflection, (c) crooked or tilted outer race and (d) crooked or tilted inner race \cite{Book}.}
\label{fig_bearing}
\end{figure}

The structure of a rolling-element bearing is illustrated in Fig. \ref{fig_bearing}, which contains the outer race typically mounted on the motor cap, the inner race to hold the motor shaft, the balls or the rolling elements, and the cage for restraining the relative distances between adjacent rolling elements \cite{Book}. The four common scenarios of misalignment that are likely to cause bearing failures are demonstrated in Fig. \ref{fig_bearing}(a) to (d).
Since bearing is the most vulnerable component in a motor drive system, accurate bearing fault diagnostics has been a research frontier for engineers and scientists for the past decades. Specifically, this problem has been approached by developing a physical model of bearing faults, and understanding the relationship between bearing faults and measurable signals, which can be captured by a variety of sensors and analyzed with signal processing techniques. Sensing modalities that have been explored include vibration \cite{VIB1, VIB2}, acoustic noise \cite{AE1, AE2}, stator current \cite{TGH1, GR1}, thermal-imaging \cite{Thermal}, and multiple sensor fusion \cite{Sensor}, among which vibration analysis is the most dominant. The existence of a bearing fault as well as its specific fault type can be readily determined by performing frequency spectral analysis on the monitored signals and analyzing their components at characteristic fault frequencies, which can be calculated by a well-defined mechanical model \cite{Book} that depends on the motor speed, the bearing geometry and the specific location of the bearing defect.  
%


However, accurately identifying the presence of a bearing fault can be challenging in practice, especially when the fault is still at its incipient stage and the signal-to-noise ratio of the monitored signal is small. In addition, unlike other motor failures (stator inter-turn, broken rotor bar, etc. \cite{TGH2}) that can be accurately determined by electric signals, the uniqueness of a bearing failure lies in its multi-physics nature. It is the primary mechanical vibration due to the bearing defect that triggered the abnormal electric signal, which further influences the output torque, the motor speed, and finally the bearing vibration pattern itself, whose fault frequency is directly proportional to the motor speed. Furthermore, the accuracy of the traditional physical model-based vibration analysis can be further affected by background noise due to external motion and vibration, and its sensitivity is also subject to change with respect to sensor mounting positions and spatial constraints in a highly-compact environment. Therefore, instead of vibration analysis, a popular alternative approach is to analyze the stator current signal \cite{TGH1, GR1}, which has already been measured in motor drives to regulate the motor's torque and speed, and thus it would not bring extra device or installation costs. 

Despite its advantages such as economic savings and simple implementation, the motor current signature analysis (MCSA) can encounter many practical issues. For example, the magnitude of stator currents at the bearing fault frequency can vary at different loads, different speeds, and different power ratings of the motors themselves, thus bringing challenges to identify a universal threshold of the stator current to trigger a fault alarm at an arbitrary operating condition. Therefore, a thorough and systematic commissioning stage is usually required while the motor is still at the healthy condition, and the healthy data would be collected while the target motor is running at different loads and speeds. However, this process, summarized as a ``Learning Stage'' in patent US5726905 \cite{Patent1}, can be tedious and expensive to perform, and needs to be repeated for any new motor with a different power rating.
%
%

Most of the challenges described above can be attributed to the fact that all of the conventional model-based methods rely solely upon the threshold value of different signals (data) at the fault frequencies to determine the presence of a bearing fault. These models can only describe the signal features of a few well-defined fault types, while in reality the naturally occurring faults are often more complicated. For example, at the early stage of a fault the signatures can be less well-defined or even not traceable by using the physical models; more than one faults can occur at the same time, which potentially modifies the fault features and creates new features due to the the coupling effect. Therefore, there may exist many unique features or patterns hidden in the data themselves that can potentially reveal a bearing fault, and it is almost impossible for humans to identify these convoluted features through manual observation or interpretation. Therefore, many researchers have applied various machine learning (ML) algorithms, including artificial neural networks (ANN), principal component analysis (PCA), support vector machines (SVM), etc., to parse the data, learn from them, and apply what they have learned to make intelligent decisions regarding the presence of bearing faults \cite{Review00, Review01, Review1, Review2}. Most of the literature applying these ML algorithms report satisfactory results with classification accuracy over 90\%.

To achieve an even better performance at versatile operating conditions and noisy environments, deep learning (DL) based methods are becoming increasingly popular to meet this demand \cite{Review3, Review4, Review5, Review6}. This literature survey incorporates more than 180 papers dedicated to bearing fault diagnosis, around 80 of which employed some type of DL approaches. The number of papers also grows exponentially over the recent years, indicating a booming interest in employing DL methods for bearing fault diagnostics.  

%

In this context, this paper seeks to present a thorough overview on the recent research work devoted to applying ML and DL techniques on bearing fault diagnostics. The rest of the paper is organized as follows. In Section II, we introduce some of the most popular datasets used for bearing fault detection. Next, in Section III, we look into some traditional ML methods, including ANN, PCA, \emph{k}-nearest neighbors (\emph{k}-NN), SVM, etc., with a brief overview of major publications applying each ML algorithm for bearing fault detection. For the main part of this paper, in Section IV, we take a deep dive into the research frontier of DL based bearing fault identification. In this section, we will provide our understanding of the research trend toward DL approaches. Specifically, we will discuss the advantages of DL based methods over the conventional ML methods in terms of fault feature extraction and classifier performance, as well as new functionalities offered by DL techniques that cannot be accomplished before. We will also provide a detailed analysis to each of the major DL techniques, including convolutional neural network (CNN), auto-encoder (AE), deep belief network (DBN), recurrent neural network (RNN), generative adversarial network (GAN), and their applications in bearing fault detection. In Section V, a comparative study is conducted on different DL algorithms to offer a more intuitive insight, which compared the classifier performance utilizing the popular open source ``Case Western Reserve University (CWRU) bearing dataset''. Finally in Section VI, detailed recommendations and suggestions are provided regrading the selection of specific DL algorithms for specific application scenarios, such as the setup environment, the data size, and the number of sensors and sensor types. Future research directions are also discussed to further improve the classifier accuracy, and facilitate domain adaptation and technology transfer from laboratories to the real-world.
\section{Popular Bearing Fault Datasets}
Data is the foundation for all of the ML methods. To develop effective ML and DL algorithms for bearing fault detection, a good collection of datasets is necessary. Since the natural bearing degradation is a gradual process and may take many years, most people conduct experiment and collect data either using bearings with artificially induced faults, or with accelerated life testing methods. While the data collection is still time consuming, fortunately a few organizations have made the effort and published their bearing fault datasets for engineers and researchers to develop their own ML algorithms. Thanks to their prevalence in the research community, these datasets can also serve as a common ground for the evaluation and comparison of different algorithms. 

Before getting into details of various ML and DL developments, in this section, we briefly introduce a few popular datasets used by most papers covered in this review.
\subsection{Case Western Reserve University (CWRU) Dataset}
\begin{figure}[!t]
\centering
\includegraphics[width=3.4in]{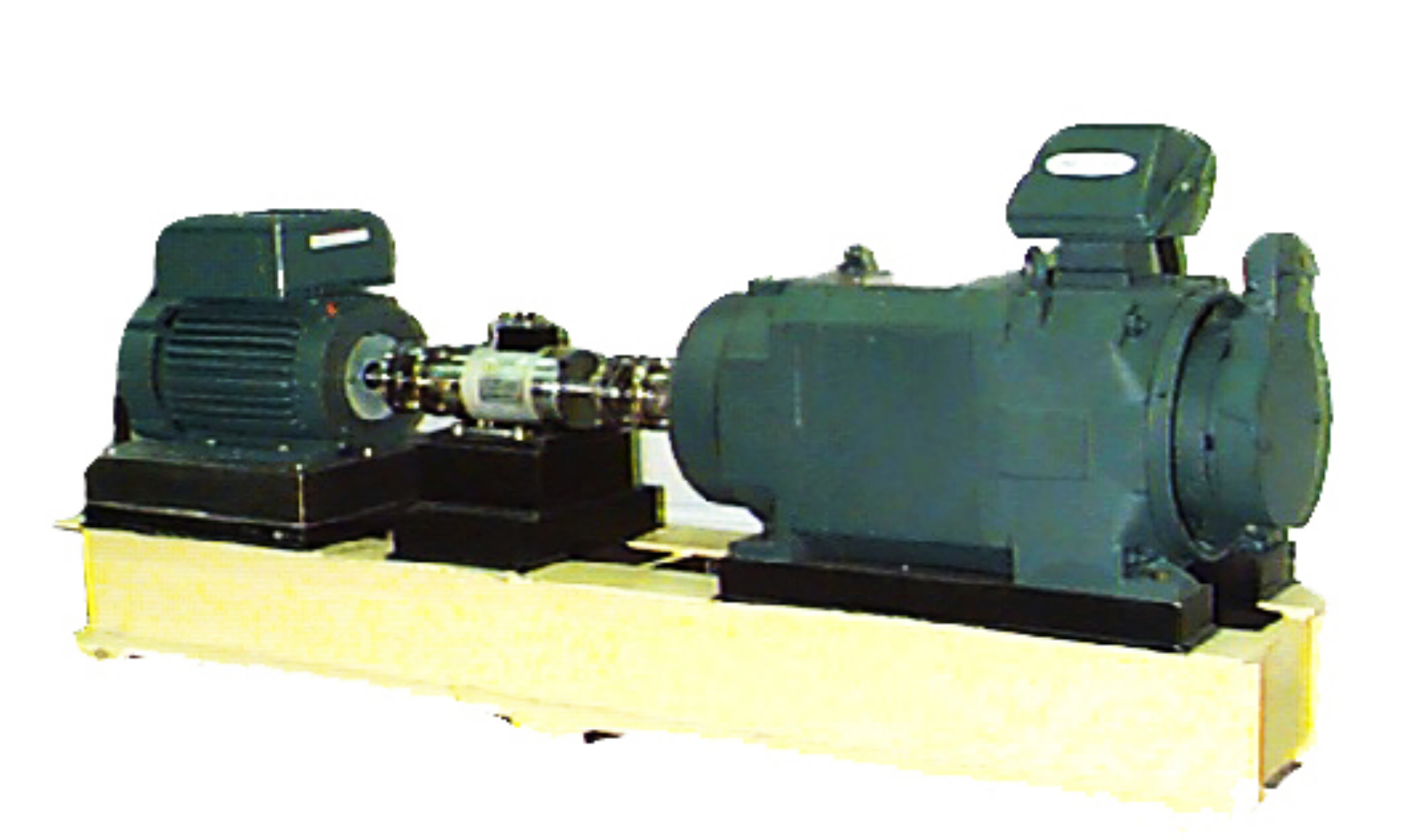}
\caption{Experimental setup for collecting the CWRU bearing dataset \cite{CWRU}.}
\label{fig_CWRU}
\end{figure}
The test stand used to acquire the Case Western Reserve University (CWRU) bearing dataset is illustrated in Fig. \ref{fig_CWRU}, in which a 2-hp induction motor is shown on the left, a torque transducer/encoder is in the middle, while a dynamometer is coupled on the right. Single point faults are introduced to the bearings under test using electro-discharge machining with fault diameters of 7 mils, 14 mils, 21 mils, 28 mils, and 40 mils, at the inner raceway, the rolling element and the outer raceway. Vibration data are collected for motor loads from 0 to 3 hp and motor speeds from 1,720 to 1,797 rpm using two accelerometers installed at both the drive end and fan end of the motor housing, and two sampling frequencies of 12 kHz and 48 kHz were used. The generated dataset is recorded and made publicly available on the CWRU bearing data center website \cite{CWRU}.

The CWRU dataset serves as a fundamental dataset to validate the performance of different ML and DL algorithms, and a comprehensive comparative study on previous work employing the CWRU dataset will be presented in Section V.
\subsection{Paderborn University Dataset}
\begin{figure}[!t]
\centering
\includegraphics[width=3.4in]{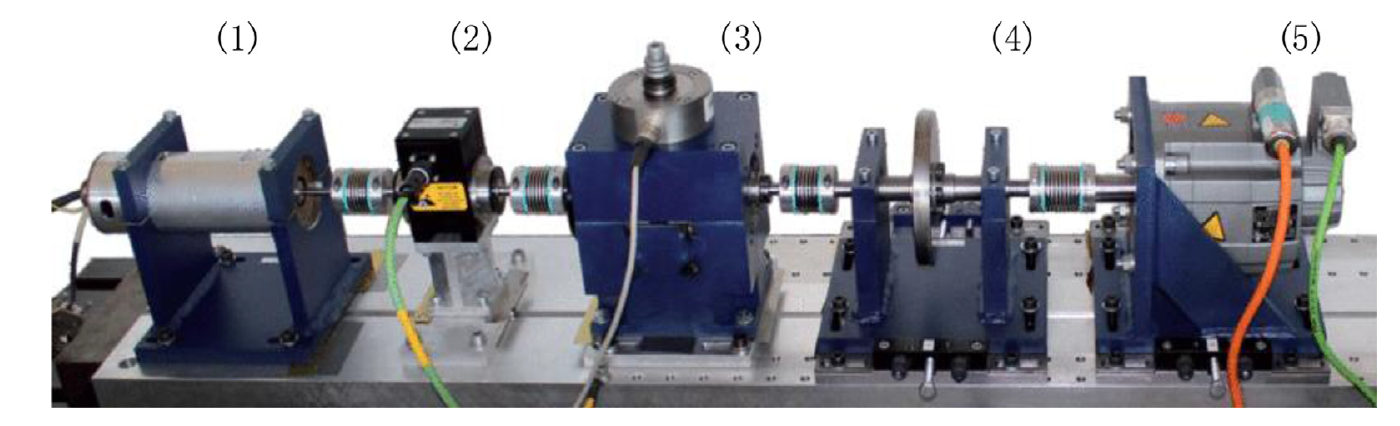}
\caption{Modular test rig collecting the Paderborn bearing dataset consisting of (1) an electric motor, 
(2) a torque-measurement shaft, (3) a rolling bearing test module, (4) a flywheel, and (5) a load motor \cite{Paderborn}.}
\label{fig_Paderborn}
\end{figure}
The Paderborn university bearing dataset \cite{Paderborn} includes the synchronous measurement of motor current and vibration signals, thus enabling the verification of multi-physics models and sensor fusion of different signals to increase the accuracy of bearing fault detection. Both stator current and vibration signals are measured with a high resolution and a high sampling rate, and experiments are performed on 26 damaged bearings and 6 undamaged (healthy) ones. Among the 26 damaged bearings, 12 are artificially damaged, and the other 14 have more realistic damages caused by accelerated life tests. This enables a more confident evaluation of ML algorithms in practical applications, where the real defects are generated through aging and the gradual loss of lubrication. The modular test rig used to acquire the Paderborn bearing dataset is illustrated in Fig. \ref{fig_Paderborn}.
\subsection{PRONOSTIA Dataset}
\begin{figure}[!t]
\centering
\includegraphics[width=3.3in]{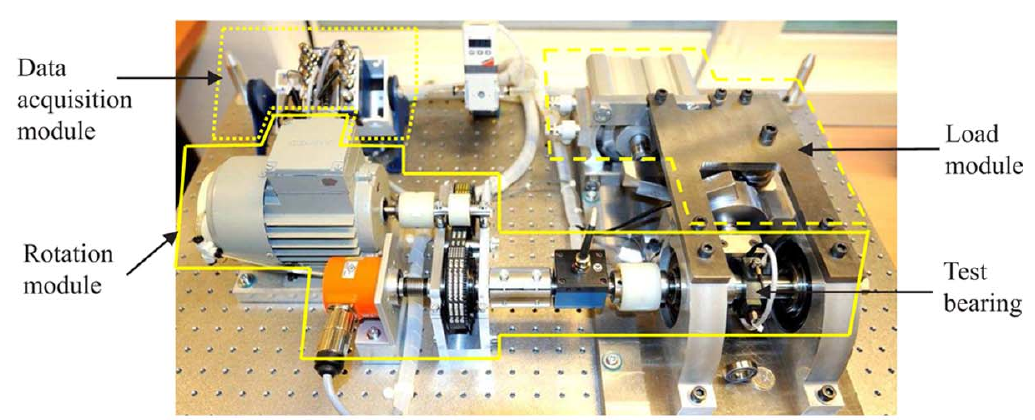}
\caption{PRONOSTIA testbed (Department of Automatic Control and Micro-Mechatronic Systems (AS2M), Franche-Comt\'e Electronique M\'ecanique
\cite{PRONOSTIA}.}
\label{fig_PRONOSTIA}
\end{figure}

Another popular dataset for predicting a bearing's remaining useful life (RUL) is known as the ``PRONOSTIA bearings accelerated life test dataset'', which serves for researchers to investigate new algorithms for bearing RUL prediction. During the International Conference on Prognostics and Health Management (PHM) in 2012, an ``IEEE PHM 2012 Prognostic Challenge'' was organized, where the PRONOSTIA degradation dataset \cite{PEM} was provided to participants allowing them to train their prognostic methods. Every participant's method was evaluated based on the estimation accuracy of the RUL of bearings under test.

The main objective of PRONOSTIA is to provide real data related to the accelerated degradation of bearings performed at varying operating conditions \cite{PRONOSTIA}. The operating conditions are characterized by two sensors: a rotating speed sensor and a force sensor. In the PRONOSTIA platform as shown in Fig. \ref{fig_PRONOSTIA}, the bearing health is monitored by gathering two types of signals: temperature and vibration (with two uni-axis accelerometers installed in the horizontal and the vertical direction respectively). Furthermore, the data are recorded with a high sampling frequency which allows the interpretation of the entire frequency spectrum of interest during the bearing degradation process. Ultimately, the monitored data can be used for post-processing to extract the relevant features offline and continuously assess the bearing's RUL.

\begin{table*}[]
\centering
\caption{Comparison of Popular Bearing Fault Datasets.}
\resizebox{\linewidth}{!}{
\begin{tabular}{lcccc}
\toprule
Dataset  & Sensor type  & Number of sensors  & Sampling frequency  & Fault mode    \\ 
\midrule
Case Western Reserve University (CWRU) Dataset & Accelerometer & 2 & 12 \& 48 kHz & Artificial  \\
\midrule
Paderborn University Dataset & \begin{tabular} [c]{@{}c@{}} Accelerometer \& current sensor \\ \& thermocouple\end{tabular} & 1 \& 2 \& 1 & 64 kHz & \begin{tabular} [c]{@{}c@{}} Artificial \\ \& accelerated aging \end{tabular}  \\
\midrule
PRONOSTIA Dataset & Accelerometer \& thermocouple & 2 \& 1 & 25.6 kHz & Natural \\
\midrule
Intelligent Maintenance Systems (IMS) Dataset & Accelerometer & 2 & 20 kHz & Natural \\
\bottomrule
\end{tabular}}
\label{tab:Dataset}
\end{table*}
\subsection{Intelligent Maintenance Systems (IMS) Dataset}
\begin{figure}[!t]
\centering
\subfloat[]{\includegraphics[width=3.4in]{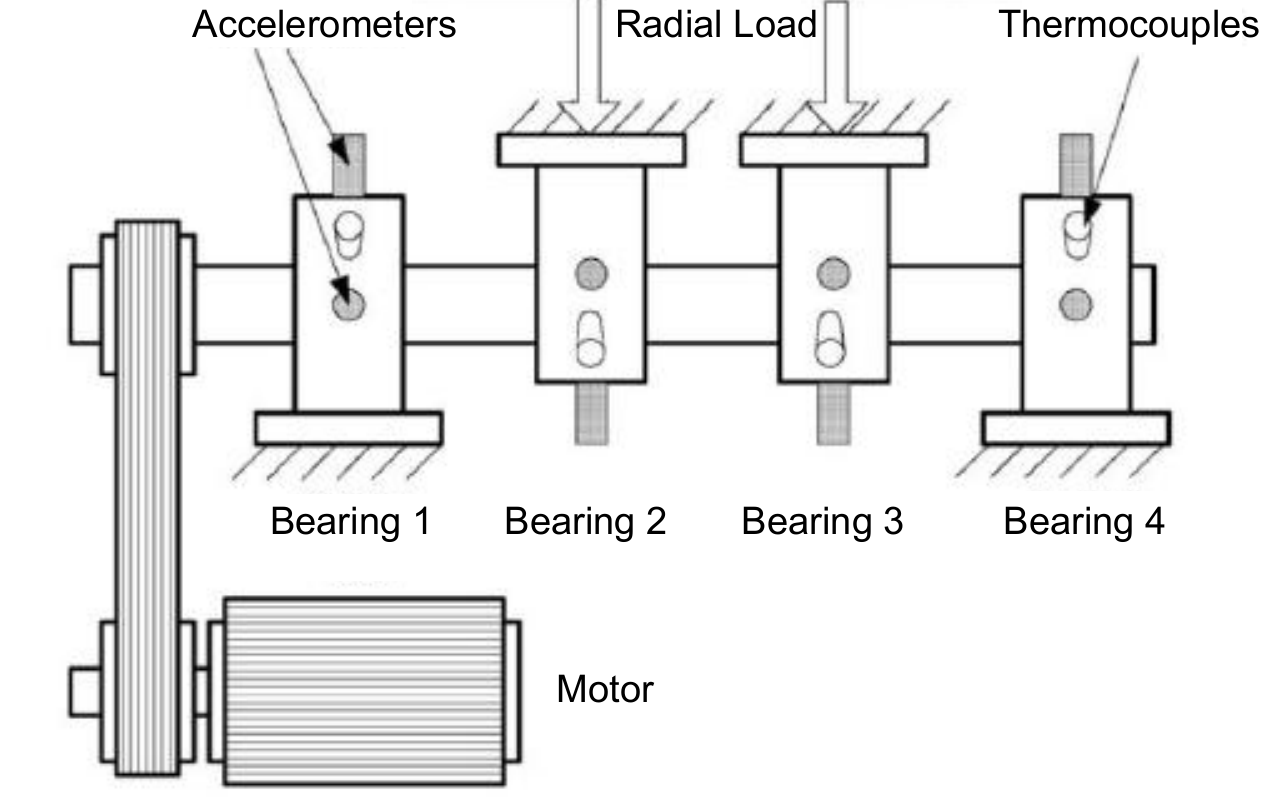}}\\
\subfloat[]{\includegraphics[width=2.6in]{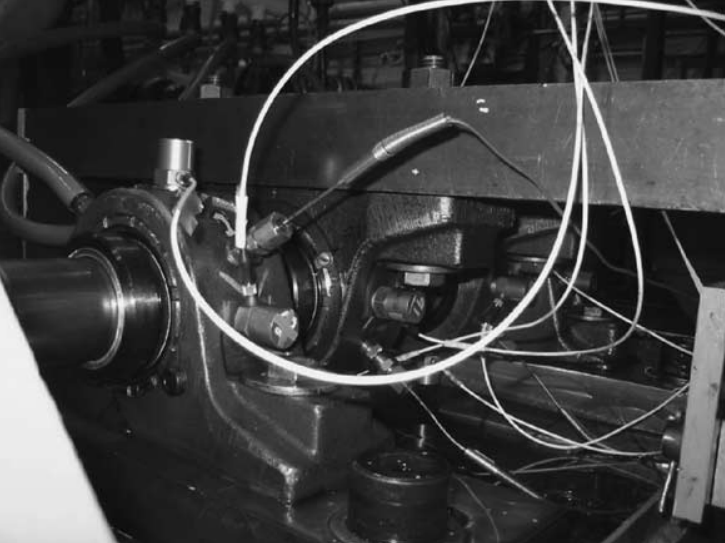}}
\caption{Illustration of the (a) bearing test rig and (b) vibration sensor placement of the IMS dataset \cite{IMS}.}
\label{fig_IMS}
\end{figure}

The IMS bearing dataset \cite{IMS_data} is generated by the NSF I/UCR Center for Intelligent Maintenance Systems (IMS) with support from Rexnord Corp. Different from the other datasets, where the bearing faults are either artificially induced by scratching or drilling the bearing surface, or created by exerting a shaft current for accelerated life testing, the IMS dataset contains a complete record of natural bearing defect evolution. Specifically, the bearing is kept running for 30 days consecutively with a constant speed of 2,000 rpm, totaling around 86.4 million cycles before a defect is confirmed \cite{IMS}. The test rig consists of four Rexnord ZA-2115 double row bearings installed on a shaft, which is coupled to an AC motor via a rubber belt 
as shown in Fig. \ref{fig_IMS}(a). A radial load of 6,000 lbs is applied onto the shaft and bearing by a spring mechanism. Two accelerometers are installed on each bearing housing, and four thermocouples are attached to the outer race of each bearing to record bearing temperature for monitoring the lubrication purposes, as shown in Fig. \ref{fig_IMS}(b).

The same experiment is repeated three times. Test 1 ends up with an inner race defect in bearing 3 and a rolling element defect in bearing 4. Test 2 and 3
end up with an outer race defect in bearing 1 and 3, respectively. The vibration data is collected every 5 or 10 minutes for a duration of 1 second with the sampling rate set at 20 kHz by National Instruments DAQCard 6062E. Since this dataset contains a complete collection of vibration signals of bearing experiments from start to failure with explicit time stamps, it is particularly suitable for predicting the RUL of rolling-element bearings. 

\subsection{Summary}
A summary of the comparing the differences between different datasets is illustrated in TABLE \ref{tab:Dataset}. So far a majority of literature on bearing fault identification with ML or DL algorithms employ the CWRU dataset due to its simplicity and popularity. The authors anticipate a growing interest on the Paderborn dataset, as it contains both the stator current signal and the vibration signal. In addition, this dataset also enables the validation of deep transfer learning and domain adaptation algorithms \cite{TL_survey, DTL_survey} to predict a more realistic bearing fault from accelerated life testing with classifiers trained on artificially induced scratches or drills. Besides, many researchers working on RUL prediction also rely on the PRONOSTIA dataset and the IMS dataset. Since the main scope of this paper is bearing fault detection, research contributions on RUL prediction is not included in this literature survey.
\section{Classical Machine Learning Based Approaches}
%
%

Before the recent  DL boom, a variety of classical ``shallow'' machine learning and data mining algorithms have been around for many years, i.e., the artificial neural network (ANN). Applying these algorithms requires a lot of domain expertise and complex feature engineering. A deep exploratory data analysis is usually performed on the dataset first, followed by dimension reduction techniques such as the principal component analysis (PCA), etc., for feature extraction. Finally, the most representative features are passed along to the ML algorithm. The knowledge base of different domains and applications can be quite different and often requires extensive specialized expertise within each field, making it difficult to perform appropriate feature extraction, or maintain a good level of transferability of ML models trained in one domain to be generalized or transferred to other contexts or settings.

Some of the earliest reviews investigating the use of artificial intelligence (AI) techniques on motor fault diagnostics can be found in \cite{Review00, Review01}, where the characteristic fault frequencies for different motor fault types are systematically summarized, and relevant papers employing ANN and fuzzy systems are discussed. In this section, a brief summary of each classical ML method will be presented, with a comprehensive list of publications for readers' reference. 
\subsection{Artificial Neural Networks (ANN)}
ANN is one of the oldest AI paradigms that has been applied to bearing fault diagnostics for almost 30 years \cite{ANN1}. In \cite{ANN1}, the bearing wear of the motor is reflected in the damping coefficient $B$ that can be inferred from a nonlinear mapping of the stator current $I$ and the rotor speed $\omega$. The complexity of obtaining an analytical expression for this nonlinear mapping is avoided by training a supervised neural network with stator current and motor speed measurements as input and predicted bearing condition as output. 35 training and 70 testing data patterns are collected on a laboratory test stand with the Dayton 6K624B-type bearing at different operating conditions. Highest bearing fault detection accuracy of 94.7\% is achieved with the conventional neural network using two input nodes $\{I,\omega\}$. The accuracy can be further improved by utilizing five input dimensions $\{I, \omega, I^2, \omega^2, I^*\omega\}$ that are manually selected. However, besides the commonly used current sensor for bearing fault diagnostics, this method requires an addition speed encoder to collect the motor speed signal as an extra input, which is not commonly available in many low-cost induction motor drives. Similarly, the rest of the papers based on ANN \cite{ANN2, ANN3, ANN4, ANN5} all require some degree of human expertise to guide its feature selection process in order to train the ANN model in a more effective manner.
\subsection{Principle Component Analysis (PCA)}
PCA is an algorithm that reveals the internal structure of the data in a way that best explains the variance in the data. If a multivariate dataset is visualized as a set of coordinates in a high-dimensional data space (one axis per variable), PCA can supply the user with a lower-dimensional projection of this object  viewed from its most informative viewpoint. Since the sensitivity of various features that are characteristics of a bearing defect may vary considerably at different operating conditions, PCA has proven itself as an effective and systematic feature selection scheme that provides guidance on manually choosing the most representative features for classification purposes. 

One of the earliest adoption of PCA on bearing fault diagnostics can be found in \cite{PCA1}. Experimental results revealed that the advantage in using only PCA identified features instead of the 13 original features is significant, as the fault diagnosis accuracy is increased from 88\% to 98\%. The study demonstrated that the proposed PCA technique is effective in classifying bearing faults with a higher accuracy and a lower number of input features when compared to using all of the original feature. Similarly, the rest of the papers based on PCA \cite{PCA2, PCA3, PCA4, PCA5 & EMD} take advantage of its data mining capability to facilitate the manual feature selection process and generate more representative features.
\subsection{K-Nearest Neighbors (k-NN)}
%
The k-NN algorithm is a non-parametric method used for either classification or regression. In \emph{k}-NN classification, the output is the class of an object, which is identified by a majority vote of its \emph{k} nearest neighbors. One early implementation of the \emph{k}-NN classifier on bearing fault diagnostics can be found in \cite{KNN1}, where \emph{k}-NN serves as the core algorithm for a data mining based ceramic bearing fault classifier based on acoustic signals. Similarly, other \emph{k}-NN based papers \cite{KNN2, KNN3, KNN&SVM} employ \emph{k}-NN to perform a distance analysis on each new data sample and determine whether it belongs to a specific fault class.
\subsection{Support Vector Machines (SVM)}
%
SVMs are supervised learning models that analyze data used for non-probabilistic classification or regression analysis. One classical work on the use of SVM towards identifying bearing faults can be found in \cite{SVM1}, where classification results obtained by the SVM are optimal in all of the cases, with an overall improvement over the performance of ANN. Other similar SVM based papers \cite{SVM2, SVM3, SVM4, SVM5, SVM6, SVM7, SVM8, SVM9, LDA&NB&SVM, SVM10, SVM11, SVM12, SVM13} also illustrated the effectiveness and efficiency of employing SVM to serve as the fault classifier.
\subsection{Others}
Besides the commonly used  ML methods listed above, many other algorithms have been applied to the identification of bearing faults, bringing in different characteristics and benefits, including neural fuzzy network \cite{NFBA, NF, MO}, Bayesian networks \cite{PCA6 & BA, BBN1, BBN2}, self-organizing maps \cite{SOM1, SOM2}, extreme learning machines (ELM) \cite{ELM1, ELM2}, transfer learning \cite{TL1, TL2, TL3}, linear discriminant analysis \cite{LD1, LD2}, quadratic discriminant analysis \cite{DC1}, random forest \cite{RF1}, independent component analysis \cite{ICA1}, softmax classifiers \cite{Softmax RE}, manifold learning \cite{Manifold1, Manifold2}, canonical variate analysis \cite{CVA}, particle filter \cite{Particle Filter 1}, nonlinear preserving projection \cite{LNPP}, artificial Hydrocarbon Networks \cite{AHN}, expectation maximization \cite{CIL}, ensemble learning \cite{EL}, multi-scale permutation entropy \cite{MPE}, empirical mode decomposition \cite{EMD1, EMD2, EMD3, EMD4, EMD5 & AdaBoost}, topic correlation analysis \cite{Topic Correlation}, affinity propagation \cite{AP}, and dictionary learning \cite{Dictionary1, Dictionary2}.
\subsection{Challenges with the Classical ML Algorithms}
As presented in the earlier sections, to detect the presence of a bearing fault using a classical ML algorithm, the characteristic fault frequencies are calculated based on the rotor mechanical speed and the specific bearing geometry, and these frequencies will serve as fault features. This feature determination process is known as ``feature engineering''. The amplitude of signals at these frequencies can be monitored to train various ML algorithms and identify any anomalies. However, such a technique may encounter many challenges that ultimately affect the classification accuracy. 
\begin{enumerate}
    \item \emph{Sliding:} The fault frequency is based on the assumption that no sliding occurs between the rolling element and the bearing raceway, i.e., these rolling elements will only roll on the raceway. Nevertheless, this is seldom the case in reality, as the rolling element often undergoes a combination of rolling and sliding movement. As a consequence, the calculated frequency may deviate from the real fault frequency and make this manually determined feature less informative of a bearing defect. 
    \item \emph{Frequency interplay:} If multiple types of bearing faults occur simultaneously, these faults will interact and the resultant characteristic frequencies can add or subtract  due to a complicated electro-mechanical process, thereby obfuscating the informative frequencies.
    \item \emph{External vibration:} There is also the possibility of interference induced from additional sources of vibration, i.e. bearing looseness and environment vibration, which can obscure the useful features. 
    \item \emph{Observability:} Some faults, such as the bearing lubrication and general roughness related faults, do not even manifest themselves as a characteristic cyclic frequency, which makes them very hard to detect with the traditional model-based spectral analysis or classical data-driven ML methods. 
    \item \emph{Sensitivity:} The sensitivity of various features that are characteristic of bearing defect may vary considerably at different operating conditions. A very thorough and systematic ``learning stage'' is typically required to test the sensitivity of these frequencies on any desirable operating condition before it can be actually put into use with the traditional approach. 
\end{enumerate}

Because of the aforementioned challenges, manually engineered features based on the bearing characteristic fault frequency can be difficult to interpret, and sometimes may even lead to inaccurate classification results, especially when applying the ``shallow'' classical ML methods that rely on human-engineered features in the training process. Therefore, many DL algorithms with automated feature extraction capabilities and better classification performance have been applied to bearing fault diagnostics, which will be discussed in detail in the next section.
\section{Deep Learning Based Approaches}
Deep learning is a subset of machine learning that achieves great power and flexibility by learning to represent the world as nested hierarchy of concepts, with each concept defined in relation to simpler concepts, and more abstract representations computed from less abstract ones. The trend of transitioning from classical ``shallow'' machine learning algorithms to deep learning can be attributed to the following reasons. 
\begin{enumerate}
\item \emph{Data explosion}: With the availability of exploding amount of data, and the application of crowdsourced labeling mechanisms such as Amazon mTurk \cite{Turk}, we are seeing a surging appearance of large scale dataset in many domains, such as ImageNet in image recognition, COCO for object segmentation and recognition, VoxCeleb in speaker identification, \emph{et al}. 
DL generally requires a large amount of labeled data. Some DL models in computer vision were trained using more than one million images. For many applications, including the diagnostics of bearing faults, such large datasets are not readily available and will be expensive and time consuming to acquire. On smaller datasets, classical ML algorithms can compete with or even outperform deep learning networks. With the increase of the amount of data, the performance of DL can significantly outperform most classical ML algorithms, as illustrated in Fig. \ref{fig_sim} \cite{Ng} by Andrew Ng.
%
\begin{figure}[!t]
\centering
\includegraphics[width=2.5in]{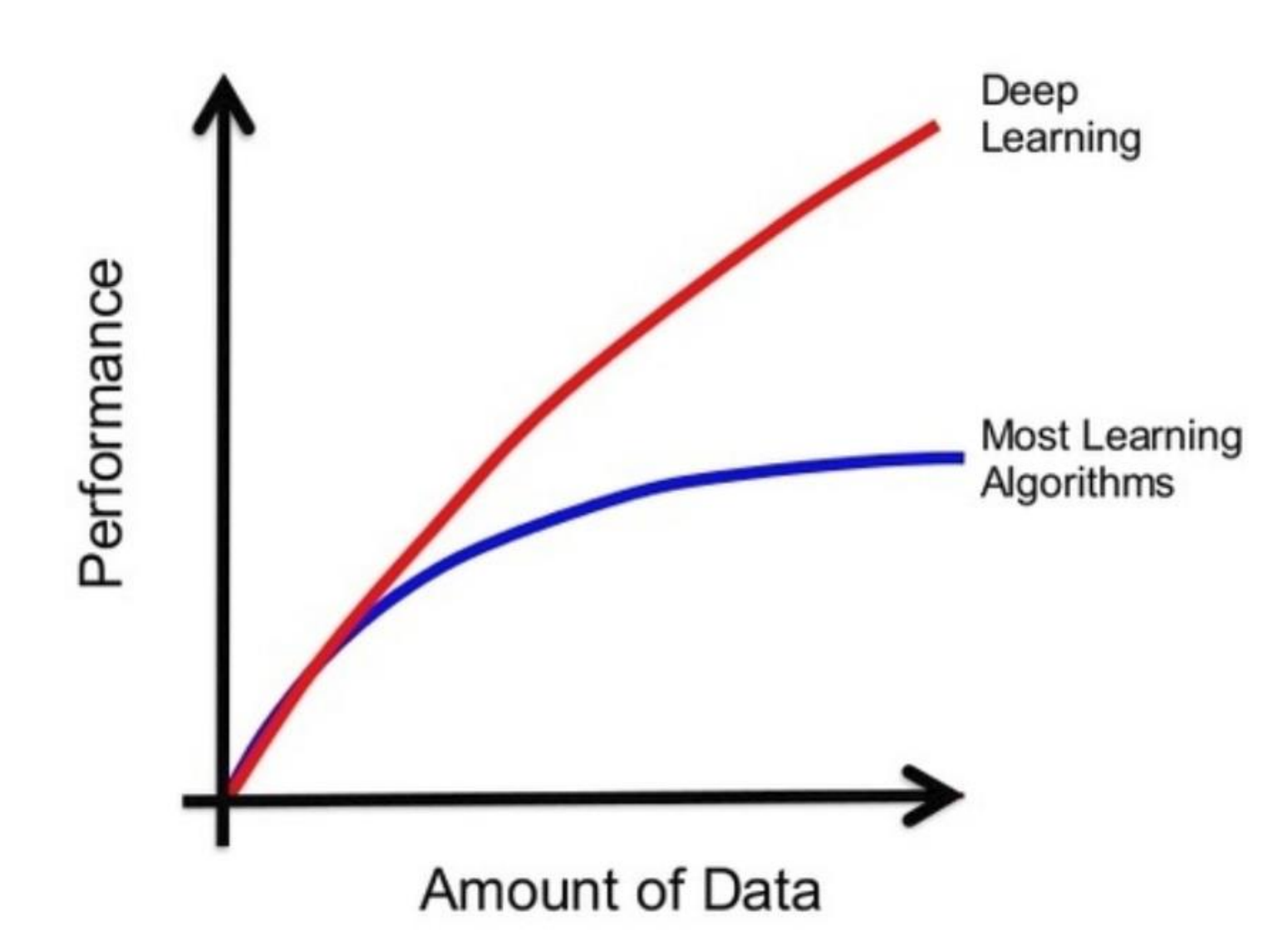}
\caption{Performance comparison of deep learning and most classical learning algorithms \cite{Ng}.}
\label{fig_sim}
\end{figure}
\item \emph{Algorithm evolution}: More techniques are being invented and getting matured in terms of controlling the training process of deeper models to achieve faster speed, better convergence, and improved generalization. For example, algorithms such as ReLU help accelerate convergence speed; techniques such as dropout and pooling help prevent overfitting; numerical optimization methods such as mini-batch gradient descent, RMSprop, and L-BFGS optimizer help leverage more data and train deeper models.
\item \emph{Hardware evolution}: Training deep networks is extremely computationally intensive, but running on a high performance GPU can significantly accelerate this training process. Specifically, GPU offers parallel computing capability and computational compatibility with deep neural networks, which makes them indispensable for training DL based algorithms. More powerful GPUs allows data scientists to quickly get the DL training up and running. For example, the NVIDIA Tesla V100 Tensor Core GPUs can now parse petabytes of data orders of magnitude faster than traditional CPUs \cite{NVIDIA}, and leverage mixed precision to accelerate DL training throughputs across every type of neural network. In the most recent years, the emergence of the accelerators for parallel computing such as GPUs, FPGAs, ASICs and TPUs have promoted the fast evolution of DL algorithms.
\end{enumerate}

All of the factors above contribute to the new era of applying DL algorithms to a variety of data-related applications. Specifically, advantages of applying DL algorithms compared to classical ML algorithms include:
\begin{enumerate}
\item \emph{Best-in-class performance}: The complexity of the computed function grows exponentially with model depth \cite{Expressivity}. DL has the best-in-class performance that significantly outperforms other solutions to problems in multiple domains, including speech, language, vision, game playing, etc.
\item \emph{Automatic feature extraction}: DL removes the need for feature engineering. Classical ML algorithms usually demand sophisticated manual feature engineering, which unavoidably requires expert domain knowledge and numerous human effort. However, when using deep neural network, there's no need for this manual process. One can simply pass the data directly to the network, and the network can automatically learn the features from raw data by auto-tuning the weights in the network. The DL network eliminates completely the challenging stage of feature engineering.
\item \emph{Transferability}: The strong expressive power and high performance of a deep neural network trained in one domain can be easily generalized or transferred to other contexts, settings or domains. Deep learning is an architecture that can be adapted to new problems relatively easily. For instance, problems in different domains such as vision, time series, and language are being solved using the same techniques like convolutional neural networks, recurrent neural networks, and long short-term memory, etc.
\end{enumerate}

Thanks to the aforementioned reasons for the transition from traditional methods to DL methods, as well as the benefits of DL algorithms discussed above, we have witnessed an exponential increase in DL applications, such as machine health monitoring and fault diagnostics, among which the bearing fault detection is a very representative case. 
\subsection{Convolutional Neural Network (CNN)}
Inspired by animal visual cortices \cite{CNN0}, the convolution operation is first introduced to detect image patterns in a hierarchical way from simple features such as edge and corner to complex features. Specifically, lower layers in the network detect fundamental lower level visual features; and layers afterward detect higher level features, which are built upon these simple lower level features. 

\begin{figure*}[!t]
\centering
\includegraphics[width=7.2in]{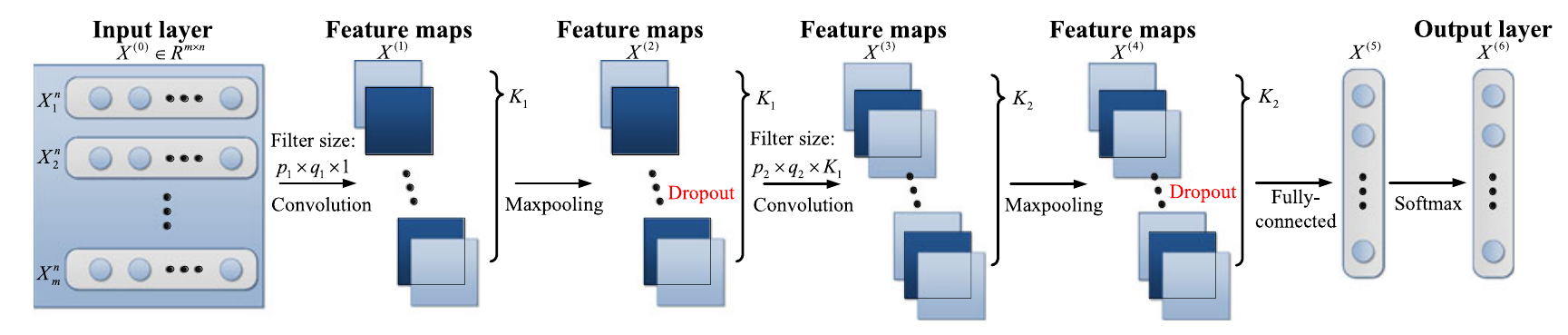}
\caption{Architecture of the CNN-based fault diagnosis model \cite{CNN6}.}
\label{fig_cnn}
\end{figure*}

The first paper employing CNN to identify bearing fault was published in 2016 \cite{CNN1}, and in the next three years many papers applying the same technique \cite{CNN2, CNN3, CNN4, CNN5, CNN6, CNN7, CNN8, CNN9, CNN10, CNN11, CNN12, CNN13, CNN14, CNN15, CNN16} have emerged and contributed to advancing bearing fault detection in various aspects. The basic architecture of a CNN-based bearing fault classifier is illustrated in Fig. \ref{fig_cnn}. Specifically, the 1-D temporal raw data obtained from different accelerometers are firstly stacked to 2-D vector form similar to the representation of images, which is then passed over to a convolutional layer for feature extraction, followed by a pooling layer for down-sampling. The combination of this convolution-pooling pattern is repeated many times to further deepen the network. Finally, the output from the hidden layers will be handed over to one or several fully-connected layers, the result of which is transferred to a top classifier based on Softmax or Sigmoid functions to determine if a bearing fault is present.

In \cite{CNN1}, the vibration data are collected using two uni-axis accelerometers installed on x- and y- direction respectively. A CNN is able to autonomously learn useful features for bearing fault detection from the raw data pre-processed by scaled discrete Fourier transform. The classification result demonstrates that the feature learning based approach significantly outperforms the feature engineering based approach of conventional ML. Moreover, another contribution of this work is to show that feature learning based approaches such as CNN can also perform bearing health prognostics, and identify some early-stage faulty conditions that have no explicit characteristic frequencies, such as lubrication degradation, which cannot be achieved using classical ML methods. 

To obtain a better trade-off between the training speed and accuracy, an adaptive CNN (ADCNN) is applied on the CWRU dataset to dynamically change the learning rate in \cite{CNN2}. The entire fault diagnosis model employs a fault pattern determination component using 1 ADCNN and a fault size evaluation component using 3 ADCNNs, and 3-layer CNNs with max pooling. Classification results demonstrate that ADCNN has a better accuracy compared to conventional shallow CNN and SVM methods, especially in terms of identifying the rolling element defect. In addition, this proposed ADCNN is also able to predict the fault size (defect width) with a satisfactory accuracy. 
On top of the conventional structure of CNN, a dislocate layer is added in \cite{CNN3} that can better extract the relationship between signals with different intervals in periodic forms, especially during the change of operating conditions. It is reported in \cite{CNN3} that the best accuracy of 96.32\% is achieved with a disclose step factor $k=3$, while the accuracy of conventional CNN without this disclose layer is only 83.39\%. 
%

Similar to earlier work \cite{CNN1, CNN2, CNN3}, \cite{CNN4} implements a 4-layer CNN structure with 2 convolutional and 2 pooling layers employing both the CWRU dataset \& dataset generated by Qian Peng Company in China, and the accuracy outperforms the conventional SVM and the shallow Softmax regression classifier, especially when the vibration signal is mixed with ambient noise. The improvement can be as large as 25\%, showcasing the excellent built-in denoising capabilities of the CNN algorithm. A sensor fusion approach is applied in \cite{CNN5}, in which both the temporal and spatial information of the CWRU raw data from two accelerometers at the drive end and the fan end are stacked by transforming 1-D time-series data into a 2-D matrix form. The average accuracy using the fusion of two sensors is increased to 99.41\% from the previous 98.35\% with only one sensor. 

Many variations of CNN are also employed to tackle the bearing fault diagnosis challenge \cite{CNN6 ,CNN7, CNN8, CNN9, CNN10, CNN11, CNN12, CNN13} using the CWRU dataset to obtain more desirable characteristics than conventional CNN. For example, a CNN based on LeNet-5 
is applied in \cite{CNN6}, which contains 2 alternating convolutional-pooling layers and a 2 fully-connected layers. Padding is used to control the size of learned features, and zero-padding is applied to prevent dimension loss. This improved CNN architecture is able to provide a better feature extraction capability with an astonishing accuracy (99.79\%) on test set, which is higher than other deep learning based methods such as the adaptive CNN (98.1\%) and the deep belief network (87.45\%). The proposed CNN based on LeNet-5 also dominates the classical ML methods such as SVM (87.45\%) and ANN (67.70\%). In addition, a deep fully convolutional neural network (DFCNN) incorporating 4 convolution-pooling layer pairs is employed in \cite{CNN7}, while the raw data are also transformed into spectrograms for easier processing. An accuracy of 99.22\% is accomplished, outperforming 94.28\% of the linear SVM with particle swarm optimization (PSO), and 91.43\% of the conventional SVM. These results are obtained using the same training set to train different networks and the same test set to evaluate and compare their performances. 

To save the extensive training time required for most CNN based algorithms, a multi-scale CNN (MS-DCNN) is adopted in \cite{CNN8}, where convolution kernels of different sizes are used to extract features of different scales in parallel. The mean accuracy of a 9-layer 1-D CNN, a 2-D CNN and the proposed MS-DCNN are 98.57\%, 98.25\% and 99.27\%, respectively. In addition to the subtle increase in accuracy compared to conventional CNNs, the number of parameters to be determined during training is only 52,172, which is significantly lower than those of 1-D CNN (171,606) and 2-D CNN (213,206). Moreover, a very deep CNN of 14 layers with training interference is used in \cite{CNN9}, which is able to maintain a high accuracy in noisy environments or during load shifts. However, the training time and the amount of parameters to be trained would increase dramatically, posing a potential threat of overfitting the data. Similarly, to overcome the impact of load variations, a novel bearing fault diagnosis algorithm based on improved Dempster-Shafer theory CNN (IDS-CNN) is employed in \cite{CNN10}. This improved D-S evidence theory is implemented via a distance matrix from the modified Gini Index. Extensive evaluations revealed that, by fusing complementary or conflicting evidences from different models and sensors, the proposed IDS-CNN algorithm is able to accommodate different load conditions and achieve a better fault diagnosis performance than  conventional DNN models and ML approaches such as SVM.

To better suppress the impact of speed variations on bearing fault diagnosis, a novel architecture based on CNN referred to as ``LiftingNet" is implemented in \cite{CNN11}, which consists of split layers, predict layers, update layers, pooling layers, and fully-connected layers, with the main learning process performed in a split-predict-update loop. A 4-class classification is carried out with the CWRU dataset randomly and evenly split into training set and test set. The final classification accuracy is 99.63\%. However, since all of the signals recorded by CWRU are measured in a small speed range (from 1,720 to 1,797 rpm), another experiment is established to record vibration signals with four distinct rotor frequencies (approximately 10, 20, 30, and 40 Hz), and the average accuracy still reaches 93.19\%, which is 14.38\% higher than conventional SVM algorithm. Similarly, a fault diagnosis method based on the Pythagorean spatial pyramid pooling (PSPP) CNN is proposed in \cite{CNN12} to enhance the classification accuracy during motor speed variations. Compared to a spatial pyramid pooling layer that has been used in an CNN, a PSPP layer is allocated in this work as a front layer of CNN, and features obtained by the PSPP layer can be delivered to convolutional layers for further feature extraction. According to the experiment result, this method has a higher diagnosis accuracy at various rotating speeds compared to other methods. In addition, the PSPP-CNN model trained by data at certain rotating speeds can be transferred and used to diagnose bearing fault at full working speed.

Since CNN excels at processing 2-D matrix data, such as images in the field of computer vision, it generally requires the transformation of 1-D time-domain vibration signal into 2-D signal to take full advantage of the strength that CNN can offer. Aiming at simplifying this conversion process and reducing the percentage of training data required due to the expensiveness of acquiring a large amount of data through experiments, an adaptive overlapping CNN (AOCNN) is proposed in \cite{CNN13} to directly process the 1-D raw vibration signal, and eliminate the shift variant problem of time-series signal. Compared to the conventional CNN, its novelty lies in the overlapping layer, which is used to sample the raw vibration signal. After the adaptive convolutional layer separates these samples into segments, sparse filtering is employed in the local layer to obtain local features. Classification results reveal that AOCNN with SF can identify ten health conditions of the bearing with a 99.19\% test accuracy when only 5\% samples are used as the training set, which is a significant improvement considering most of the DL based methods demand a minimum of 25\% data allocated in the training set. In addition, the test accuracy can further rise to 99.61\% when the test set data percentage increases from 5\% to 20\%.

Besides the CWRU dataset which contains only vibration signal, the Paderborn University bearing dataset \cite{Paderborn}, as stated in Section II, includes  synchronized stator current and vibration signals. In addition, the Paderborn dataset also incorporates both  artificially induced bearing fault and  realistic damages caused by accelerated lifetime tests. In \cite{CNN14}, the Paderborn dataset is used to train a deep inception net with atrous convolution, which improves the average accuracy from 75\% (best result of conventional data-driven methods) to 95\% for diagnosing the real bearing faults when trained only with the data generated from artificial bearing damages. The ``PRONOSTIA bearings accelerated lifetime test dataset'' \cite{PEM}, as introduced in Section II, is applied in \cite{CNN15} with a deep convolution structure consisting of 8 layers: 2 convolutional, 2 pooling, 1 flat, and 3 nonlinear transformation layers. Health indicators (HI) are later defined based on the CNN output, and the classification result shows the accuracy of HI predicted using CNN is superior than that of self-organizing maps (SOM). 
%
%

In addition to identifying damages on rolling element bearings, the adoption of CNN on spindle bearings is also discussed in \cite{CNN16}, in which the wavelet packet energy of the vibration signal is taken as input.
\subsection{Auto-encoders}
\begin{figure}[!t]
\centering
\includegraphics[width=3.45in]{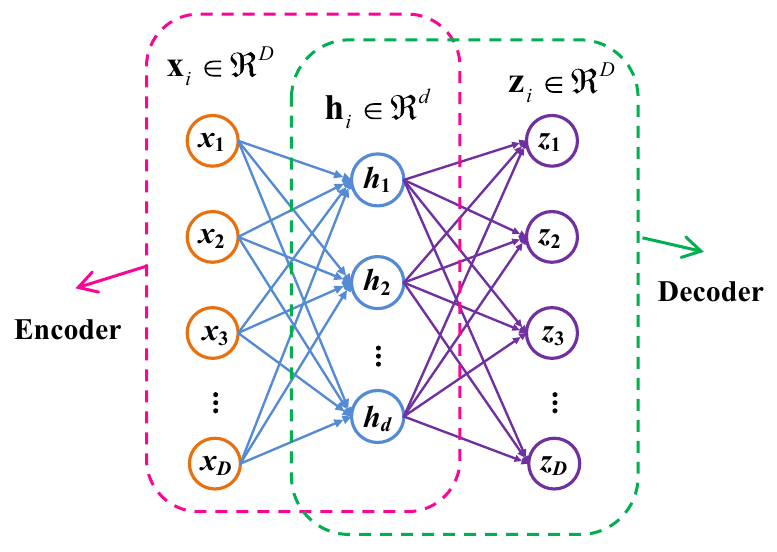}
\caption{Process of training a one hidden layer auto-encoder \cite{SA4}.}
\label{fig_ae}
\end{figure}

Auto-encoder is proposed in the 1980s as an unsupervised pre-training method for ANNs \cite{SCHMIDHUBER201585, Ballard}. After decades of evolution, the auto-encoder has become widely adopted as an unsupervised feature learning method and a greedy layer-wise neural network pre-training method. 
%
%
The training process of an auto-encoder with 1 hidden layer is illustrated in Fig. \ref{fig_ae} \cite{SA4}. Specifically, an auto-encoder is trained from an ANN, which consists of two parts: the encoder and the decoder. The output of the encoder is fed into the decoder as input. The ANN takes the mean squared error between the original input and output as the loss function, which essentially aims at generating the final output by imitating the input. After this ANN is trained, the decoder part is dropped while only the encoder part is kept. Therefore, the output of the encoder is the feature representation that can be employed in the next-stage classifier.

Among a large number of studies of applying auto-encoders to bearing fault diagnosis \cite{SA0, SA1, SA2, SA3, SA4, SA6, SA7, SA8, SA9, SA10, SA11, SA12, SA13}, an early attempt can be found in \cite{SA0}, where a 5-layer auto-encoder based DNN is utilized to adaptively extract fault features from the frequency spectrum and effectively classify the bearing health condition. The classification accuracy reaches 99.6\%, which is significantly higher than the  70\% of back-propagation based neural networks (BPNN). In \cite{SA1}, an auto-encoder based extreme learning machine (ELM) is employed, seeking to integrate the automatic feature extraction capability of auto-encoders and the high training speed of ELMs. The average accuracy of 99.83\% compares favorably against other traditional ML methods, including wavelet package decomposition based SVM (WPD-SVM) (94.17\%), EMD-SVM (82.83\%), WPD-ELM (86.75\%) and EMD-ELM (81.55\%). More importantly, the required training time drops by around 60\% to 70\% using the same training and test data, thanks to the adoption of ELM.

Compared to CNN, the denoising capability of conventional auto-encoders is not prominent. Thus in \cite{SA2}, a stacked denoising auto-encoder (SDA) is implemented, which is suitable for deep architecture based robust feature extraction on signals containing ambient noise under varying working conditions. This specific SDA consists of three auto-encoders stacked together. To strike a balance between classification performance and training speed, three hidden layers with 100, 50, and 25 units respectively are employed. The original CWRU bearing data are perturbed by a 15 dB random noise to manually create a noisy background, and data from multiple operating conditions are used as the test set to evaluate its denoising capability at different speeds and loads. The average classification result reveals the proposed SDA is able to achieve a worst case accuracy of 91.79\%, which is 3\% to 10\% higher when compared to the conventional 
SAE without the denoising capability, and classical ML algorithms such as SVM and random forest (RF). Similar to \cite{SA2}, another form of SDA is utilized in \cite{SA3} with three hidden layers of (500, 500, 500) units. Signals from the CWRU dataset are mixed with different levels of artificially induced noise in the time domain, and later transformed to the frequency domain. The proposed method has a better diagnosis accuracy than deep belief networks, particularly with the added noises, where an average improvement of 7\% is achieved.

In \cite{SA4}, a locomotive bearing dataset developed at the Northwestern Polytechnical University 
is used to validate the performance of auto-encoders. Based on this dataset, the authors adopted the maximum correntropy as the loss function instead of the traditional mean squared error, and an artificial fish-swarm algorithm (AFSA) is used to optimize the key parameters in the loss function of the auto-encoder. Results show that the customized 5-layer auto-encoder composed of this maximum correntropy loss function and AFSA algorithm outperforms standard auto-encoder by an accuracy of 10\% to 40\% in a 5-class classification problem. Similarly, a new deep AE constructed with DAE and contractive auto-encoder (CAE) is applied to the locomotive bearing dataset for enhancing the feature learning capability. A single DAE is firstly used to learn low-layer features from the raw vibration data, then multiple CAEs are used to learn deeper features. In addition, locality preserving projection (LPP) is also adopted to fuse these deep features to further improve the quality of the learned features. The classification accuracy of this mixed DAE-CAE-LPP approach is 91.90\%, showcasing the advantage over the standard DAE (84.60\%), the standard CAE (85.10\%), and the classical ML algorithms of BPNN (49.70\%) and SVM (57.60\%). However, all of the auto-encoder based methods are also 6 to 10 times more time-consuming when compared to classical ML methods. 

In addition. an aircraft-engine inter-shaft bearing vibration dataset with the inner race, the outer race, and the rolling element defect is adopted as the input data in \cite{SA6}, where a new AE based on Gaussian radial basis kernel function is employed to enhance the feature learning capability. Later, a stacked AE is developed using this new AE and multiple conventional AEs. An average accuracy of 86.75\% is achieved, which is much better compared to the standard SAE (44.90\%) and the standard DBN (19.65\%). Moreover, the importance of the proposed Gaussian radial basis kernel function is showcased in a comparative study. When the Gaussian kernel function is changed to a polynomial kernel function (PK) and a power exponent kernel function (PEK), the accuracy would drop to 24.25\% and 65.55\%, respectively.

Similar to the case of CNN, many variations of SAE are also employed in the last two years to tackle the bearing fault diagnosis problem \cite{SA7, SA8, SA9, SA10, SA11, SA12, SA13} using the popular CWRU dataset, and all of which have achieved some form of performance elevation when compared to the traditional SAE. In \cite{SA7}, an ensemble deep auto-encoder consisting of a series of auto-encoders (AE) based on different activation functions is proposed for unsupervised feature learning from the measured vibration signal. Later, a decision ensemble strategy is designed to merge the classification result from each individual AE and ensure an accurate and stable diagnosis result. An average classification accuracy of 99.15\% is achieved, which performs better than many classical ML methods including BPNN (88.22\%), SVM (90.81\%), and RF (92.07\%) based on a manually selected feature of 24 dimensions. Similarly, by altering the activation function, a deep wavelet auto-encoder (DWAE) with extreme learning machine (ELM) is implemented in \cite{SA8}, where the wavelet function is employed as the nonlinear activation function, enabling wavelet auto-encoders (WAE) to effectively capture signal characteristics. Then a DWAE with multiple WAEs is constructed to enhance the unsupervised feature learning ability, and ELM is adopted as the output classifier. Based on the final result, this method (95.20\%) not only outperforms the classical ML methods such as BPNN (85.43\%) and SVM (87.97\%), but also some standard DL algorithms, including the standard DAE with Softmax (89.70\%) and the standard DAE with ELM (89.93\%). 

Considering the relatively large data size required to train deep neural nets, a 4-layer DNN with stacked sparse auto-encoder is established in \cite{SA9} with a compression ratio of 70\%, indicating only 30\% of the original data are needed to train the proposed model. The DNN has 720 input nodes, 200 and 60 nodes in the first and the second hidden layer, and 7 nodes in the output layer, the number of which depends on the number of fault conditions. A nonlinear projection is performed to compress the vibration data and perform adaptive feature extraction in the transformed space, and the accuracy of the proposed method reaches 97.47\%, which is 8\% higher than the SVM, 60\% higher than a three-layer ANN, and 46\% higher than a multi-layer ANN. \cite{SA10} summarizes two limitations of the conventional SAE. Firstly, an SAE tends to extract similar or redundant features that increase the complexity rather than the accuracy of the model. Secondly, the learned features may have shift variant properties. To overcome these issues, a new SAE-LCN (local connection network) is proposed, which consists of the input layer, the local layer, the feature layer, and the output layer. Specifically, this method learns features from the input signal locally in the local layer, then obtains shift-invariant features in the feature layer, and finally recognizes the bearing health condition in the output layer for a 10-class classification problem. The average accuracy is reported to reach 99.92\%, which is 1\% to 5\% higher than EMD, ensemble NN, and DL based methods. Similarly, a diagnosis model using SAE and incremental support vector machines is implemented in \cite{SA11}, which is tested for online diagnosis purposes.

Besides the most commonly used Softmax classifiers in the output layer, the Gath-Geva (GG) clustering algorithm is implemented in \cite{SA12}, which induces a fuzzy maximum likelihood estimation (FMLE) of the distance norm to determine the likelihood of a sample belonging to each cluster. While an 8-layer SDAE is still used to extract the useful features from the vibration signal, GG is deployed to identify the different fault types. The worst case classification accuracy is 93.3\%, outperforming the classical EMD based feature extraction schemes by almost 10\%.

To further reduce the DL based model complexity, another bearing fault diagnosis method based on a fully-connected winner-take-all auto-encoder is proposed in \cite{SA13}, in which the model explicitly imposes lifetime sparsity on the encoded features by keeping only k\% largest activations of each neuron for all of the samples in a mini-batch. A soft voting method is implemented to increase the classification accuracy and stability by aggregating the prediction result of each signal segment sliced by a sliding window. A customized dataset is generated to test the diagnosis performance under a noisy environment by adding white Gaussian noise to the original CWRU dataset. The experimental result demonstrates that with a simple \emph{two-layer} network, the proposed method not only handles the bearing fault detection with a higher precision at normal conditions, but  also demonstrates a better anti-noise capability when compared to some deeper and more complex models, such as a deep CNN.
\begin{figure}[!t]
\centering
\includegraphics[width=2.6in]{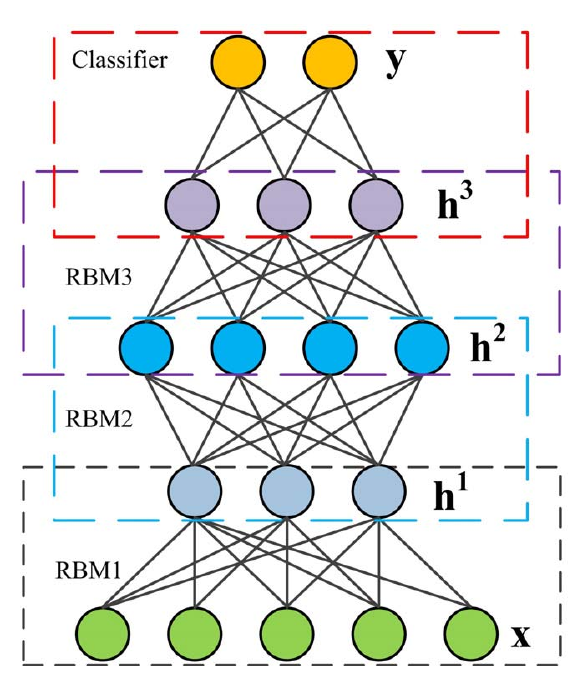}
\caption{Architecture of DBN \cite{DBN1 & SA}.}
\label{fig_dbn}
\end{figure}
\begin{figure*}[!t]
\centering
\includegraphics[width=4.6in]{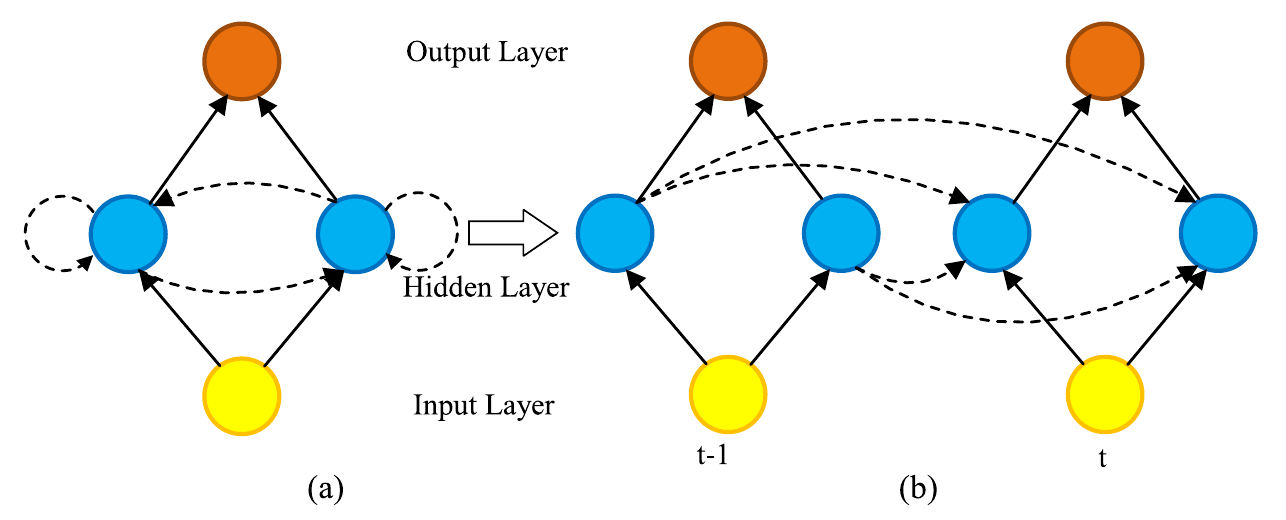}
\caption{Architecture of (a) RNN, and (b) RNN over a time step \cite{RNN1}.}
\label{fig_rnn}
\end{figure*}
\subsection{Deep Belief Network (DBN)}
In DL, a deep belief network (DBN) can be viewed as a composition of simple unsupervised networks such as restricted Boltzmann machines or auto-encoders, where each sub-network's hidden layer serves as the visible layer for the next, as illustrated by different colored boxes in Fig. \ref{fig_dbn}. An RBM is an undirected generative energy-based model with a ``visible'' input layer, a hidden layer, and connections in between, but not within layers. This composition leads to a fast layer-by-layer unsupervised training procedure, where contrastive divergence is applied to each sub-network in turn, starting from the ``lowest'' pair of layers in the architecture.

This greedy layer-by-layer training process has led to one of the first effective DL algorithms \cite{DBN0}. There are many attractive implementations of DBNs in real-life applications such as natural language understanding and drug discovery; and its first application on bearing fault diagnosis was published in 2017 \cite{DBN1 & SA}. 

In \cite{DBN1 & SA}, a multi-sensor vibration data fusion technique is implemented to fuse the time-domain and frequency-domain features extracted using multiple 2-layer SAEs. Then a 3-layer DBN is used for classification purposes. Validation is performed on the vibration data collected at different speeds, and the 97.82\% accuracy demonstrates that the proposed method can effectively identify a bearing fault at multiple operating conditions. The feature visualization using t-SNE reveals that this multi-SAE based feature fusion outperforms other methods with only one SAE or without fusion. In \cite{DBN2 & CNN}, a stochastic convolutional DBN is implemented by means of stochastic kernels and averaging, and an unsupervised CNN is built to extract 47 features. Later a 2-layer DBN is implemented with (28, 14) nodes, 5 kernels in each layer, and 1 pooling layer without overlapping. Finally, a Softmax layer is used for classification, and the average accuracy exceeds 95\%.

Many DBN papers also take the CWRU bearing dataset as the input data \cite{DBN3, DBN4, DBN5} due to its popularity. For example, an adaptive DBN and dual-tree complex wavelet packet (DTCWPT) is proposed in \cite{DBN3}. The DTCWPT first prepossesses the vibration signal to generate a feature set with 9$\times$8 feature parameters. Then a 3-level wavelet decomposition of the signal is performed using the order 5 Daubechies wavelet as the basis function. 
Then a 5-layer adaptive DBN of the (72, 400, 250, 100, 16) structure is used for bearing fault classification. The average accuracy is 94.38\%, which is much higher compared to the classical ML methods such as ANN (63.13\%), GRNN (69.38\%), and SVM (66.88\%) using the same training and test data. In \cite{DBN5}, data from two accelerometers mounted on the load end and fan end respectively are processed by multiple DBNs for feature extraction; then faulty conditions based on the extracted features are determined by Softmax; and the final health condition is fused by D-S evidence theory. An accuracy of 98.8\% is accomplished considering the load variation from 1 to 3 hp, a significant improvement when compared to the conventional SAE and CNN. Similar to this D-S theory based output fusion method \cite{DBN4}, a 4-layer DBN of the (400, 200, 100, 10) structure with different hyper-parameters coupled with ensemble learning is implemented in \cite{DBN5}. Specifically, an improved ensemble method is used to acquire the weight matrix for each DBN, and the final diagnosis result is formulated from each DBN based on their weights. The average accuracy of 96.95\% is better than that of a single DBN of different weights (mostly around 80\%), as well as a simple voting ensemble scheme based DBN (91.21\%). 

Besides the CWRU bearing dataset, many other datasets have been used to evaluate the performance of DBN on bearing fault diagnostics. In \cite{DBN6}, a convolutional DBN constructed with convolutional RBMs is applied on the locomotive bearing vibration, where an auto-encoder is firstly used to compress the data and reduce its dimension. Without any feature extraction process, the compressed data are divided into training samples and test samples to be fed into the convolutional DBN. The convolutional DBN based on Gaussian visible units is able to learn the representative features, overcoming the problem of conventional RBMs that all visible units must be related to all hidden units by different weights. Lastly, a Softmax layer is used for classification and obtains an accuracy of 97.44\%, which compares favorably against other DL methods, such as the denoising auto-encoder (90.76\%), the standard DBN (88.10\%), and the standard CNN (91.24\%), using the same classifier and raw data. In \cite{DBN7}, a bearing dataset directly obtained from power plants is used to evaluate the performance of a 5-hidden-layer DBN with (512, 2048, 1024, 2048, 512) nodes in each layer. The dataset contains vibration signals collected from various scaled applications, such as small testbeds and real field deployments. The unsupervised feature extraction is performed by DBN, and the fault classifier is designed using SOM which achieves a 97.13\% accuracy. 

DBN has also been applied to bearing RUL prediction. In \cite{DBN8}, a DBN-feed-forward neural network (FNN) is applied to perform automatic feature learning with DBN and RUL prediction with FNN. Two accelerometers are 
%
%
%
mounted on the bearing housing, in directions perpendicular to the shaft, and the data is collected with a 102.4 kHz sampling frequency for a duration of 2 seconds. Experimental results demonstrate the proposed DBN based approach can accurately predict the true RUL as the bearing approaches the point of failure, and the accuracy of the predictions tends to increase and converge over time.

\subsection{Recurrent Neural Network (RNN)}
Different from a feed-forward neural network, a recurrent neural network (RNN) processes the input data in a recurrent behavior, and its architecture is shown in Fig. \ref{fig_rnn}. With a flow path going from the hidden layer to itself, when unrolled in sequence, it can be viewed as a feed-forward neural network in the input sequence. As a sequential model, it can capture and model sequential relationships in sequential data or time-series data. However, often trained with back-propagation through time, RNN has the notorious gradient vanishing/exploding issue stemmed from its nature. Although the RNN is proposed as early as the 1980s, it has limited applications due to this reason, until the birth of long short-term memory (LSTM) in 1997. Specifically, LSTM is augmented by adding recurrent gates called ``forget'' gates. Designed for overcoming the gradient vanishing/exploding issue, LSTM has shown an astonishing capability in memorizing and modeling the long-term dependency in data, and therefore taken a dominant role in time-series and textual data analysis. So far, it has received great successes in the field of speech recognition, handwriting recognition, natural language processing, video analysis, etc. 

One of the earliest applications of RNN on bearing fault diagnostics is reported in 2015 \cite{RNN1}, where fault features are firstly extracted using the discrete wavelet transform and later selected based on the orthogonal fuzzy neighbourhood discriminative analysis. These features are then fed into an RNN to perform bearing fault detection. The experimental result has shown that
the proposed scheme based on RNN is capable of accurately detecting and classifying the bearing fault. Another RNN based health indicator (RNN-HI) is proposed in \cite{RNN2} to predict the RUL of bearings with LSTM cells used in RNN layers. Along with time-frequency features, the related-similarity (RS) feature calculates the similarity between the currently monitored data and the data at an initial operation point. After performing a correlation and monotonicity-metrics-based feature selection process, the selected features are transferred to an RNN network to predict the bearing HI, from which the RUL can be estimated. With the input dataset collected from generator bearings of wind turbines, the proposed RNN-HI is demonstrated to offer better performance than an SOM based method.

In addition, a methodology of a combined 1-D CNN and LSTM to classify bearing fault types is presented in \cite{RNN3}, where the entire architecture is composed of a 1-D CNN layer, a max pooling layer, a LSTM layer, and a Softmax layer as the top classifier. The system input is the raw signal without any pre-processing, and the best test accuracy of different configurations reaches 99.6\%. A more recent work employing a deep recurrent neural network (DRNN) is proposed in \cite{RNN4} with stacked recurrent hidden layers and LSTM units. A loss function with mean squared errors is introduced and the stochastic gradient descent (SGD) method is used as the optimizer. Besides, an adaptive learning rate is also adopted to improve the training performance. The average accuracy on the test set using the proposed method is 94.75\% and 96.53\% at 1,750 and 1,797 rpm respectively.
\begin{figure*}[!t]
\centering
\includegraphics[width=5.0in]{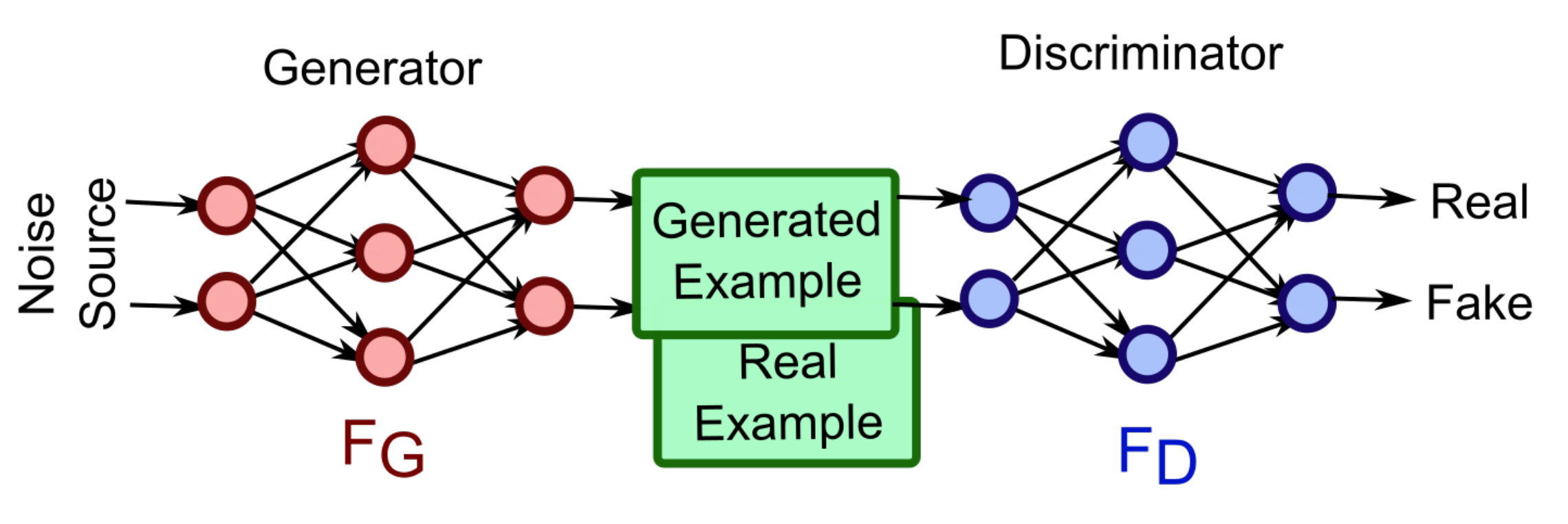}
\caption{Architecture of GAN \cite{CMU}.}
\label{fig_gan}
\end{figure*}
%
%




    


\subsection{Generative Adversarial Network (GAN)}
%

Generative Adversarial Network (GAN) was proposed by Goodfellow \emph{et al}. \cite{GANOrigin} in 2014 and rapidly became one of the most exciting breakthroughs in the field of deep learning. A GAN is composed of two parts: the generator $F_G$ and the discriminator $F_D$, as illustrated in Fig. \ref{fig_gan} \cite{CMU}. The two parts are competing with each other in a way that the generator $F_G$ is trying to confuse the discriminator $F_D$, and $F_D$ is trying to distinguish samples generated by $F_G$ from samples in the original dataset. Established as a zero-sum game framework, both $F_G$ and $F_D$ are competing to obtain an increasingly stronger capability of imitating the original data samples and discriminating in an iterative manner. 

A GAN is mainly designed for generative purposes to generate samples or functions as generation modules. Despite its relatively short history, GAN has been rapidly applied to the field of bearing fault diagnostics. One of the earliest publications appears in 2017 \cite{GAN0}, which aims at addressing the class imbalance issue using GAN. In addition, GAN is also combined with the adaptive synthetic sampling (ADASYN) approach to 
achieve meaningful oversampling when the original data samples are sparse. Comparison against standard oversampling techniques shows the superiority of adopting GAN. In \cite{GAN4}, A novel approach for fault diagnosis based on deep convolution GAN (DCGAN) with imbalanced dataset is proposed. A new DCGAN model \cite{ref_DCGAN} with 4 convolutional layers serving as the discriminator and the generator is designed and applied on raw and imbalanced vibration signals. After performing data balancing using the DCGAN model, statistical features based on time-domain and frequency-domain data are extracted to train a SVM classifier for bearing fault classification. Both the training and the test accuracy of the proposed DCGAN method demonstrate better performance than other class balancing methods, including random over-sampling, random under-sampling, and synthetic minority over-sampling technique. 

We can find a number of research works in the field of bearing fault diagnostics  employing GAN and its variants for data augmentation purposes due to their excellent generative capability. Besides that, there are also some works using GAN as the main framework to realize classification tasks, which heavily rely on the assumption that the data structure in latent space, although without labels, contains information that can be used to infer the labels. When a GAN is learning from unlabeled samples in a unsupervised manner, it can additionally learn the data distribution in latent space that distinguishes the data's unknown classes. In this way, the discriminator of GAN can be refined as a classifier assisted by some other modules in the framework. This class of GAN-centered frameworks has shown superiority in semi-supervised areas, especially in applications where labeled data are expensive and scarce.

In \cite{GAN1}, for example, the authors proposed a novel GAN framework referred to as the categorical adversarial auto-encoder (CatAAE), which automatically trains an auto-encoder through an adversarial training process, and imposes a prior distribution on the latent coding space. In the next step, a classifier tries to cluster the input examples by balancing the mutual information between examples and their predicted categorical class distributions. The latent coding space and the training process are presented to investigate the advantage of the proposed model. Experiments at different signal-to-noise ratios (SNRs) and different motor load levels have indicated the preponderance of the proposed CatAAE in learning useful characteristics when compared to the categorical generative adversarial networks (CatGAN) and the K-means algorithm.

Since many real-world applications do not comply with the common assumption that the training set and the test set have the same distribution, due to the fact that the operating condition may vary frequently. Similar to \cite{GAN1} and inspired by GAN, a new adversarial adaptive 1-D CNN model (A2CNN) is proposed in \cite{GAN2} to address this problem. Experiments show that the A2CNN has a strong fault-discriminative and domain invariant capacity, and therefore its prediction can achieve a high accuracy even at different operating conditions. Other works employing GAN to tackle the data imbalance issue can be found in \cite{GAN3} and \cite{GAN5}.
\begin{figure*}[!t]
\centering
\includegraphics[width=5.6in]{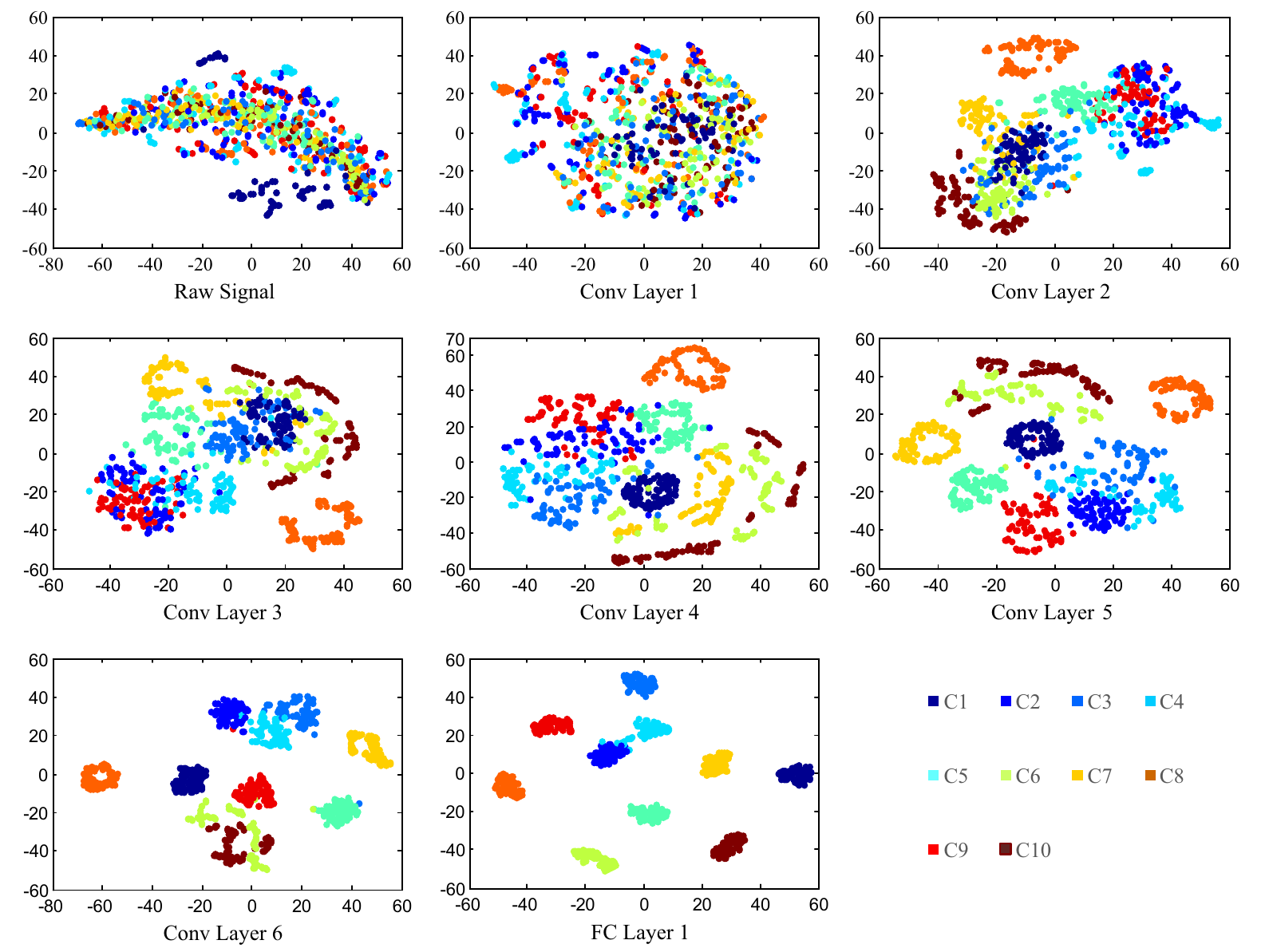}
\caption{Feature visualization via t-SNE: feature representations for all test signals extracted from raw signal, six convolutional layers and the last fully connected layer respectively \cite{CNN9}.}
\label{fig_visual}
\end{figure*}
\begin{table*}[ht!]
\centering
    \caption{A Summary of Different Deep Learning Architecture.}
\begin{tabular}{c|m{5cm}|m{5cm}}
\toprule
    \textbf{Architecture} & \hspace{0.65in} \textbf{Description} & \hspace{0.65in} \textbf{Characteristics} \\ 
    \midrule
    \begin{minipage}{.35\textwidth}
      \includegraphics[width=1\columnwidth]{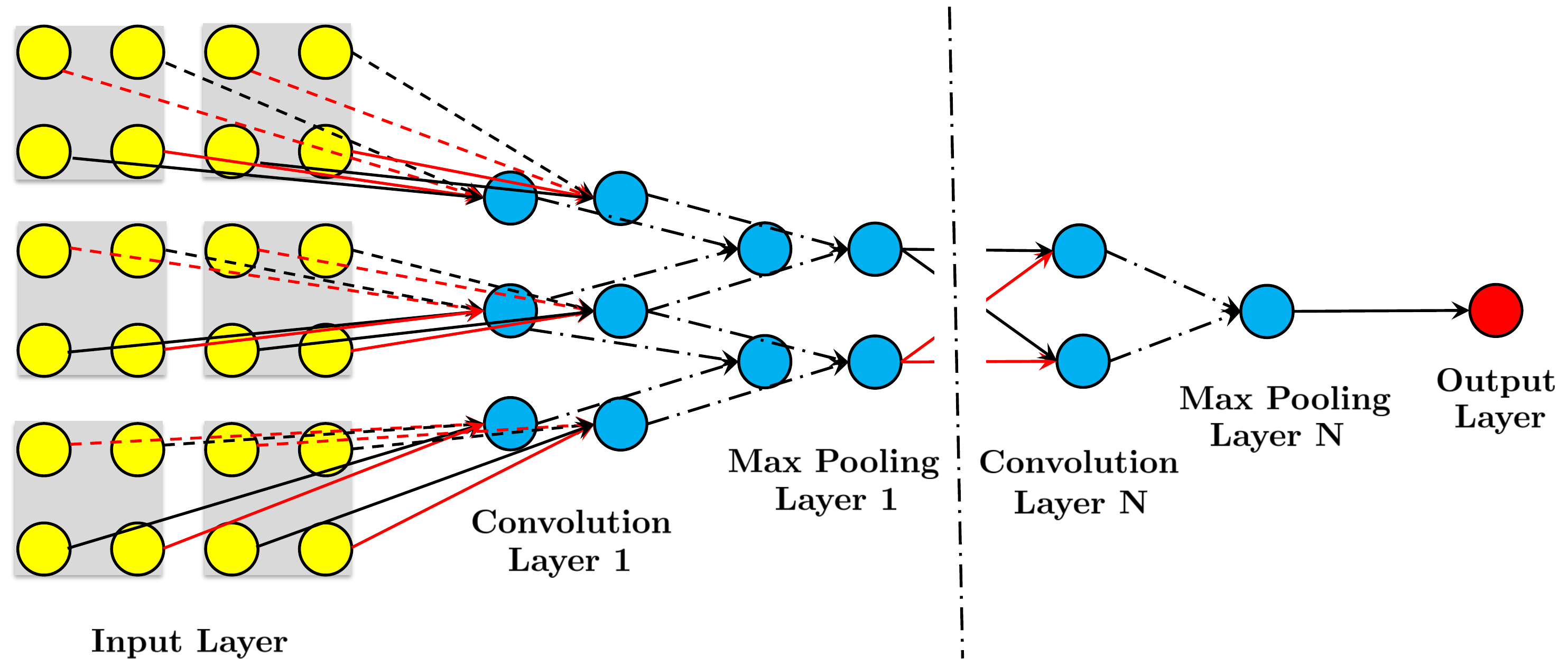}
    \end{minipage}
    &
    \hspace{0.3in} \textbf{Convolutional Neural Network}
    \begin{itemize}
        \vspace{0.05in}
        \vspace{0.05in}
        \item Well-suited for 2-D data, i.e., images, thus 1-D temporal data need to be pre-processed to form 2-D vectors
        \vspace{0.05in}
        \item ReLU after convolutional layers helps accelerate the convergence speed. 
        \vspace{0.05in}
        \item Many variants have been proposed: ADCNN \cite{CNN2}, LiftingNet \cite{CNN11}, and inception net \cite{CNN14}, etc. 
        \vspace{0.05in}
    \end{itemize}
    & 
    \vspace{0.08in}
    \textbf{Pros:}
      \begin{itemize}
        \item Few neuron connections required with respect to a typical ANN.
        \item The classical CNN exhibits a good denosing capability \cite{CNN4}. 
      \end{itemize}
    \textbf{Cons:}
      \begin{itemize}
        \item May require many layers to find an entire hierarchy.
        \item May require a large labeled dataset.
      \end{itemize}
\\ 
\midrule
    \begin{minipage}{.35\textwidth}
    \vspace{0.02in}
    \hspace{0.05\columnwidth}
      \includegraphics[width=0.82\columnwidth]{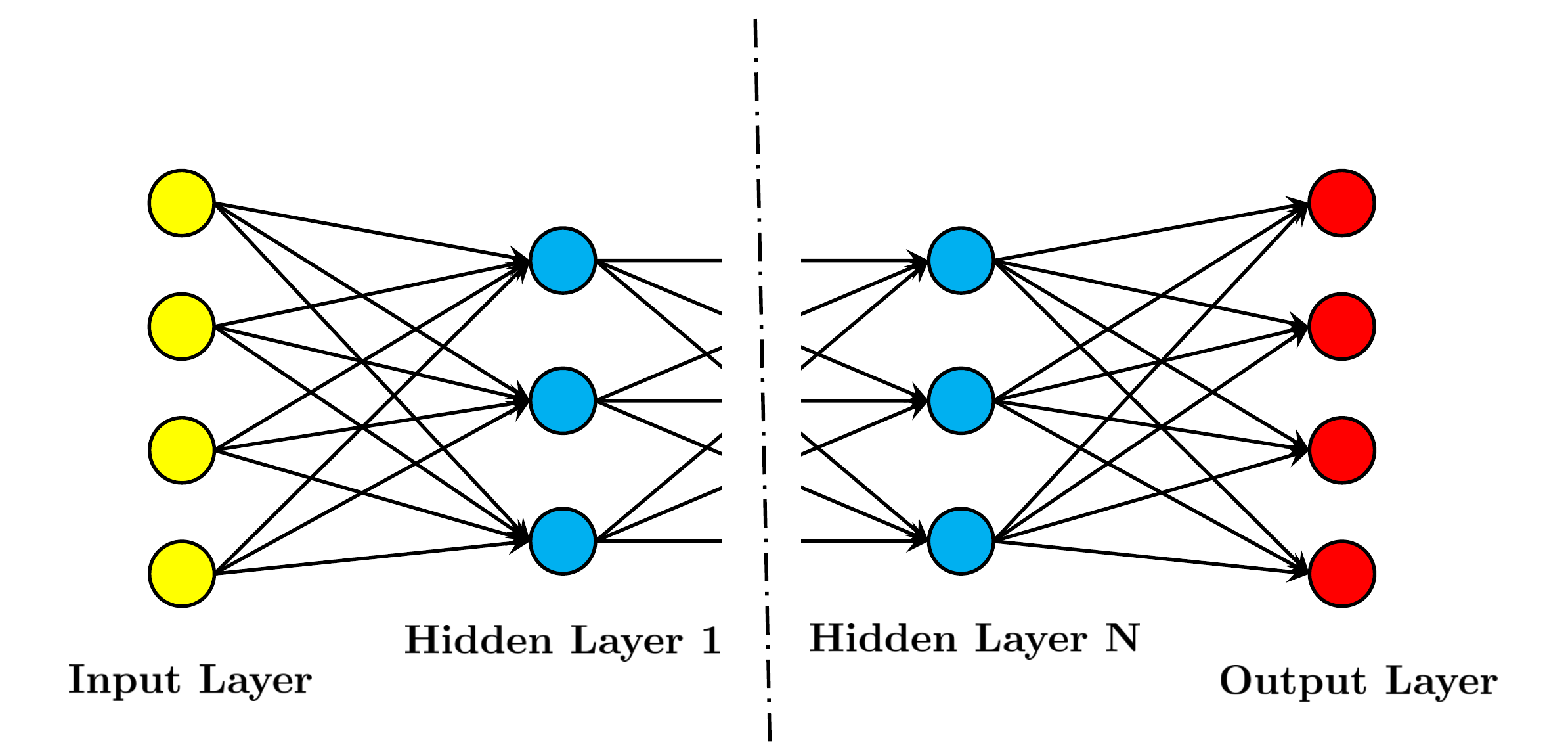}
    \end{minipage}
    &
    \hspace{0.55in} \textbf{Deep Autoencoder}
    \begin{itemize}
        \vspace{0.05in}
        \item Mainly designed for feature extraction or dimension reduction.
        \vspace{0.05in}
        \item Unsupervised learning method aiming at reconstructing the input vector.
         \vspace{0.05in}
        \item Many variants have been proposed: stacked denoising AE \cite{SA2}, deep ensemble AE \cite{SA7}, and stacked sparse AE \cite{SA9}.
    \end{itemize}
    & 
    \vspace{0.08in}
    \textbf{Pros:}
      \begin{itemize}
        \item Does not require labeled data.
        \item Many AE variants can make the algorithm more noise-resilient and robust.
      \end{itemize}
    \textbf{Cons:}
      \begin{itemize}
        \item Requires a pre-training stage.
        \item Training may suffer from the vanishing of errors.
      \end{itemize}
\\
\midrule
    \begin{minipage}{.35\textwidth}
    \vspace{0.02in}
    \hspace{0.02\columnwidth}
      \includegraphics[width=0.94\columnwidth]{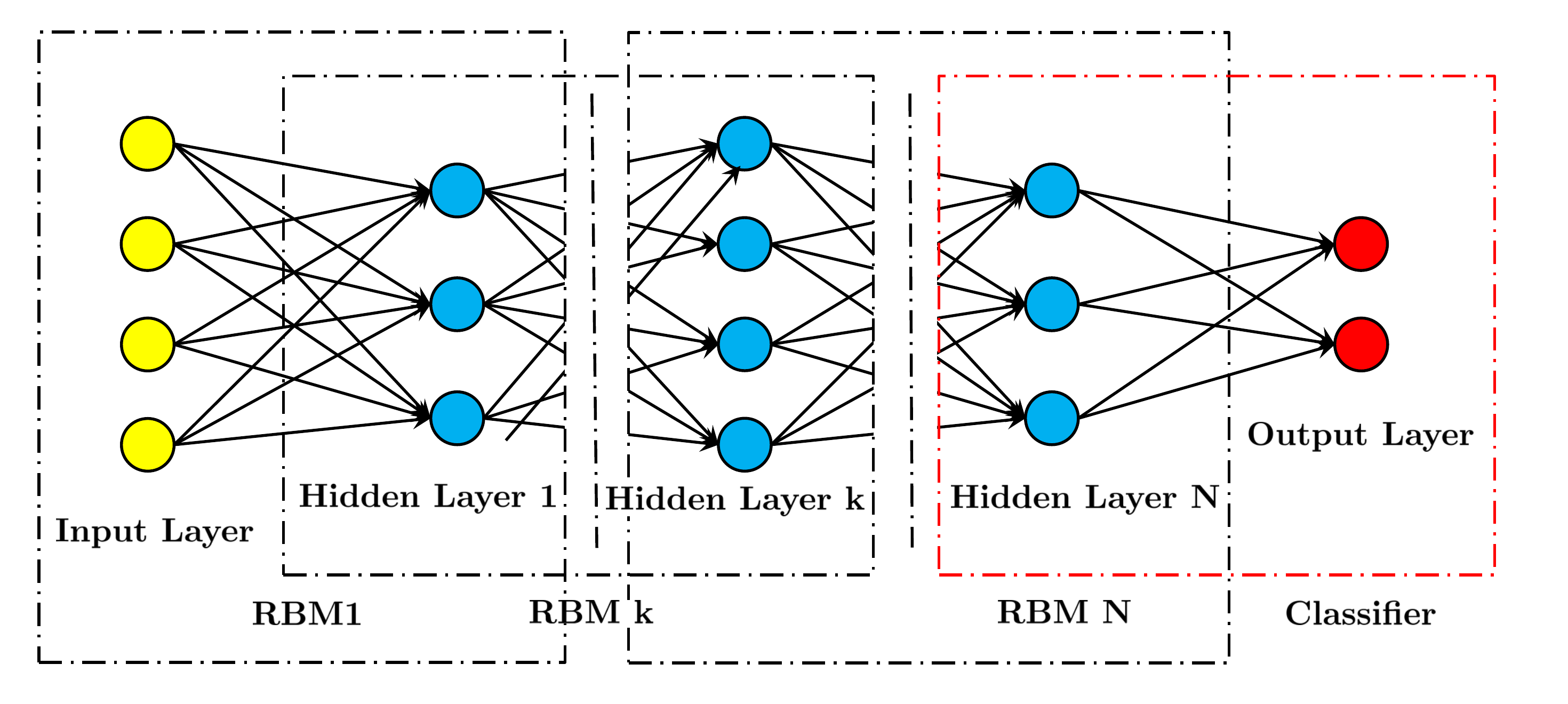}
    \end{minipage}
    &
      \hspace{0.5in} \textbf{Deep Belief Network}
      \begin{itemize}
        \vspace{0.05in}
        \item Composed of RBMs where each sub-network's hidden layer serves as the visible layer for the next.
        \vspace{0.05in}
        \item Has undirected connections just at the top two layers.
        \vspace{0.05in}
        \item Allows unsupervised and supervised training of the network.
      \end{itemize}
    & 
    \vspace{0.08in}
    \textbf{Pros:}
      \begin{itemize}
        \item Proposes a layer-by-layer greedy learning strategy to initialize the network.
        \vspace{0.05in}
        \item Tractable inferences maximize the likelihood directly.
      \end{itemize}
    \textbf{Cons:}
      \begin{itemize}
        \item Training may be computationally expensive due to the initialization process and sampling stage.
      \end{itemize}
\\
\midrule
    \begin{minipage}{.35\textwidth}
    \vspace{0.02in}
    \hspace{0.05\columnwidth}
      \includegraphics[width=0.96\columnwidth]{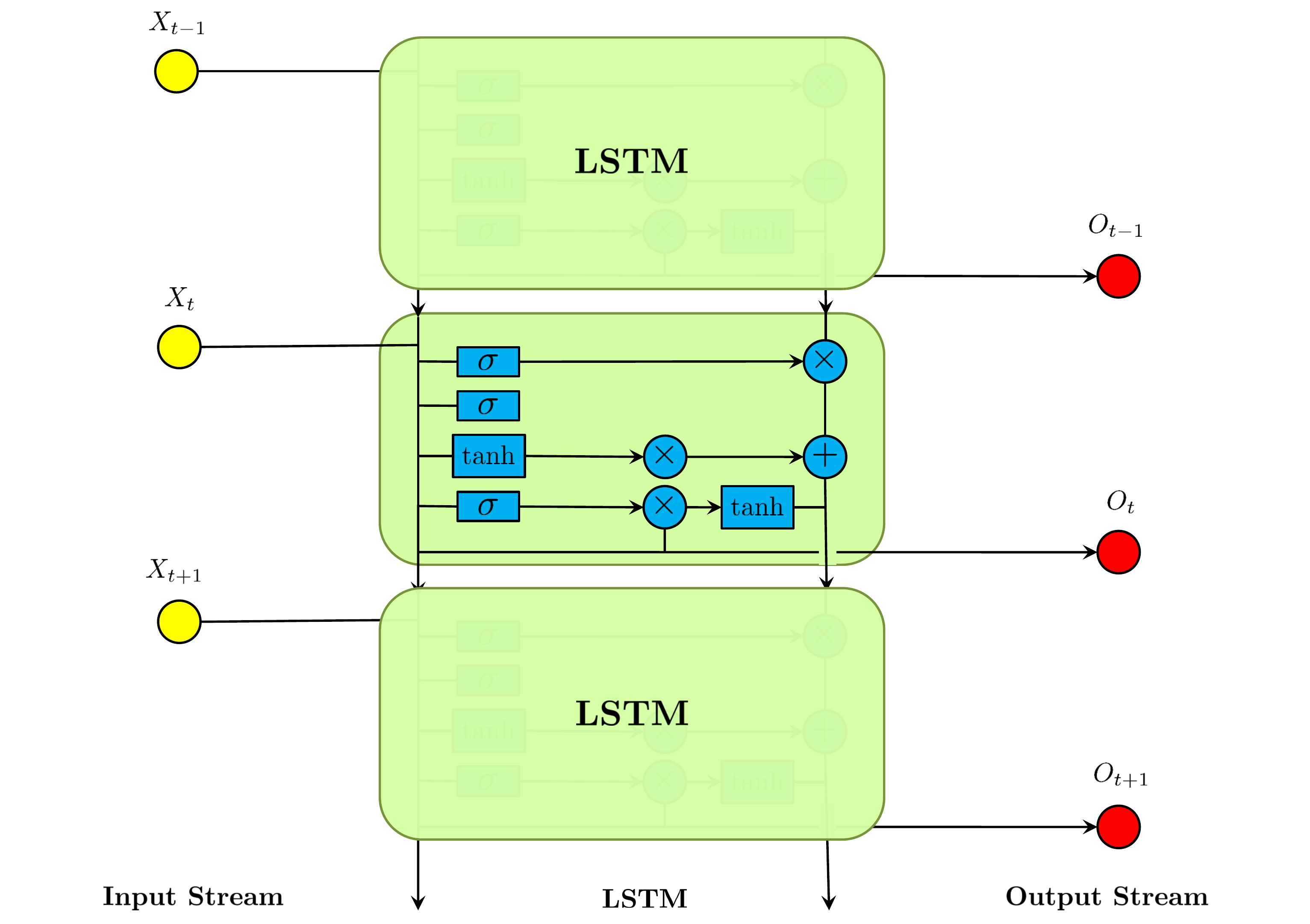}
    \end{minipage}
    &
      \hspace{0.35in} \textbf{Recurrent Neural Network}
      \begin{itemize}
        \vspace{0.05in}
        \item An ANN capable of analyzing 1-D sequential or temporal data streams.
        \vspace{0.00in}
        \item LSTM re-vibrated the application of RNNs.
        \vspace{0.05in}
        \item Suitable for applications where the output depends on the previous computations.
      \end{itemize}
    & 
    \textbf{Pros:}
      \begin{itemize}
        \item Memorizes sequential events.
        \vspace{0.05in}
        \item Capable of modeling time dependencies.
        \vspace{0.05in}
        \item Capable of receiving inputs of variable lengths.
      \end{itemize}
    \textbf{Cons:}
      \begin{itemize}
        \item Frequent learning issues due to gradient vanishing/exploding.
      \end{itemize}
\\
\midrule
    \begin{minipage}{.35\textwidth}
    \vspace{0.02in}
    \hspace{0.02\columnwidth}
      \includegraphics[width=1\columnwidth]{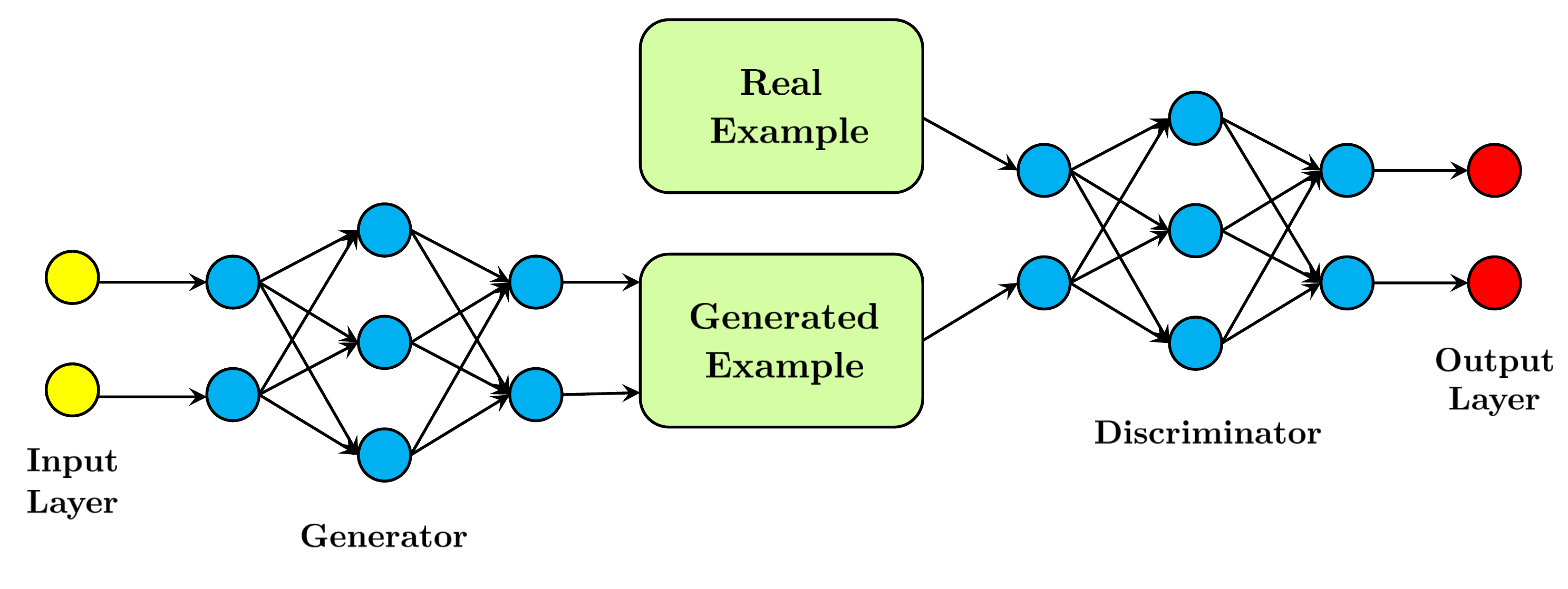}
    \end{minipage}
    &
    \vspace{0.12in}
    \hspace{0.16in} \textbf{Generative Adversarial Network}
    \begin{itemize}
        \vspace{0.05in}
        \item Composes of a generator and a discriminator. Originally designed to generate images that imitate real photos. 
        \vspace{0.05in}
        \item Applied for data augmentation in labeled data scarce applications
        \vspace{0.05in}
        \item Also used in classification tasks, usually in with semi-supervised manner.
    \end{itemize}
    & 
    \vspace{0.08in}
    \textbf{Pros:}
      \begin{itemize}
        \item Requires almost no modifications when transferring to new applications.
        \item Requires no Monte Carlo approximations to train.
        \vspace{0.05in}
        \item Does not introduce deterministic bias.
      \end{itemize}
    \textbf{Cons:}
      \begin{itemize}
        \item GAN training is unstable as it requires finding the Nash equilibrium of a game.
        \vspace{0.05in}
        \item Hard to learn to generate discrete data, such as text.
      \end{itemize}
\\
\bottomrule
\end{tabular}
\label{tab:DL_sum1}
\end{table*}

\subsection{Deep Learning based Transfer Learning}
The success of ML and DL based bearing fault diagnostics relies on a massive amount of heavily annotated data. However, this is generally not feasible in most real-world applications due to 1) the dangerous and serious consequences when machines are running at faulty conditions; 2) the potential time-consuming degradation process before the desired failure appears; and 3) the possibility of a large number of operating conditions with different speeds and loads. With DL methods trained with either the publicly available datasets or self-collected datasets sampled in a laboratory environment, the classification accuracy will naturally deteriorate when determining the presence of bearing fault in a real-world application. Even if the data collected from the same machines and bearings are used, certain level of distribution discrepancy inevitably exists between features of the training and  test sets if they were at different loads or speeds. As a result, the performance still suffers. 

Designed to tackle this practical and widely existing issue in numerous applications, transfer learning has aroused extensive attention in the machine learning community, and various transfer learning frameworks are proposed based on classical ML algorithms \cite{TL_survey, DTL_survey, TL_base0, TL_base1}. A popular method among all types of transfer learning approaches is domain adaptation. By exploring domain-invariant features, domain adaptation establishes the knowledge transfer from the source domain to the target domain \cite{DNN_TL1}. Therefore, with labeled data from the source domain and unlabeled data from the target domain, the distribution discrepancy between the two domains can be mitigated by domain adaptation algorithms. Over the last few years, an integration of deep learning and transfer learning approaches has been prevalent. Specially designed domain adaptation modules are combined with deep learning architectures to endow the domain transfer ability while maintaining the extraordinary automatic feature learning ability \cite{DNN_TL1, CNN_TL0, CNN_TL1, CNN_TL2, GAN_TL1}. 

Specifically, a domain adaptation module is proposed in \cite{CNN_TL1} to facilitate a 1-D CNN to learn domain-invariant features by maximizing the domain recognition error and minimizing the probability distribution distance. To validate the efficacy of domain adaptation, 3 datasets including CWRU dataset, IMS dataset, and railway locomotive bearing dataset are employed. By training on one of the three datasets and testing on another one, an average accuracy of 86.3\% is achieved, which has surpassed the conventional CNN of 53.1\%, and two existing domain adaptation frameworks of 75.6\% \cite{TL_base0} and 78.8\% \cite{TL_base1}.

A novel framework WDCNN (deep CNN with wide first-layer kernels) combined with adaptive batch normalization (AdaBN) was proposed in \cite{CNN_TL2}. Taking the raw vibration signal as input, the new framework is based on a CNN architecture with wide kernels (64) in the first convolutional layer to better suppress the high frequency noise. Then domain adaptation is implemented by extracting the mean and variance of the target domain signals and passing them to AdaBN. The CWRU dataset is used to conduct cross-domain experiments by training the proposed WDCNN in one working condition and testing in another one. An average accuracy of 90.0\% is achieved, which is further improved to 95.9\% by mixing with AdaBN, outperforming the conventional FFT-DNN method of 78.1\%. When tested in a noisy environment (with additive white Gaussian noise), WDCNN with AdaBN achieves a 92.65\% accuracy under a -4 dB SNR, in comparison to 66.95\% without AdaBN.

Deep generative network can also be combined with domain adaptation to produce a novel framework. In \cite{GAN_TL1}, a 2-stage structure is proposed. In the first stage, an 8-layer CNN component including 3 convolutional layers with basic classifiers is trained as the feature extractor to optimize the classification error under source supervision. Then $N_c-1$ ($N_c$ is the number of classes) CNN components, each of which consists of 3 convolutional layers and 3 dropout layers, are trained to minimize the maximum mean discrepancy. In the second stage, with the feature extractor trained in the first stage, a cross-domain classifier is trained to generate the final diagnosis result.

\subsection{Other Variants}

There are also many other DL variants implemented to cope with some of the open issues in the field of bearing fault diagnostics. Some of the selected variants are summarized as follows.
\subsubsection{Variational Autoencoders}
Proposed by Kingma \emph{et al.} \cite{VAE}, the variational auto-encoder (VAE) is different from other autoencoder variants in that it uses the variational inference to generate a latent representation of the data, and impose a distribution over the latent variables and the data itself. Compared to some general class DL algorithms, the implementation of VAE in bearing fault detection is a relatively new domain. A representative work of such is presented in \cite{VAE_1}, where a fully unsupervised deep VAE-based approach is proposed to tackle the high dimensionality of data used for failure diagnosis. Specifically, the VAE is able to extract discriminative features from the high-dimensional input data to form their corresponding low-dimensional latent space representations. The experimental results show that the VAE is a more competent and promising tool for dimensionality reduction than PCA. Besides discriminative feature learning, it is also worthwhile exploring its generative capabilities for fault diagnosis in the context of semi-supervised learning.
\subsubsection{Capsule Neural Networks}
The capsule network is a new deep learning architecture proposed by Hinton \emph{et al.} in \cite{CapsuleOrigin}, which has a strong capability to identify the position and orientation relationship of features through the capsule module. Meanwhile, its  relatively simple structure with a limited number of parameters significantly promotes the model generalization.\\
\\
For instance, in \cite{Capsule_1}, a capsule network with inception block is proposed aiming at improving the model generalization capability. This goal is achieved by employing an inception-module-augmented capsule network to adopt different working conditions, upon taking the two-dimensional short-time Fourier transform graph of the raw data as input. Besides, a regression branch is added to predict the size of the bearing defect. The experiments showcase better domain adaptation performance than state-of-the-art algorithms when training and testing on bearing fault data collected at different work loads. The authors in \cite{Capsule_2} also proposes a deep capsule network with stochastic delta rule (DCN-SDR) for bearing fault diagnosis under varying working conditions and noisy environment. The network first receives raw temporal signal as input, and then  extracts noise-immune representative features via incorporating a noise injection module, which is a regularization method based on SDR. The superiority of the proposed architecture is verified through extensive experiments and the subsequent feature visualization via t-SNE. Similarly, a capsule network combined with the Xception module (XCN), an extreme version of inception module, is developed in \cite{Capsule_3}, aiming at improving the classification accuracy of the proposed variant of the capsule network. Trained on  ideal laboratory conditions and tested on an actual system setup, the proposed diagnostic model delivers improved classification accuracy, robustness, and training speed. 
\subsubsection{Siamese Neural Networks}
Originally proposed by Bromley and LeCun \cite{Siamese_origin} in the early 1990s, the siamese neural network is designed to solve signature verification as an image matching problem. A siamese neural network consists of twin networks, which compares distinct inputs and rank similarities between them. With a growing interest in few-shot learning over the recent years due to insufficient data, Koch et al. \cite{Siamese} implemented the siamese neural network for one-shot image recognition, which inspired the later work of applying this similar siamese network structure to bearing fault diagnostics \cite{Few-Shot}. Specifically, in the specific siamese network model in \cite{Few-Shot}, two identical networks are set up to take in sample pairs of the same or different categories, which can measure the distance of the two feature vector outputs to determine their similarity. Compared to the WDCNN benchmark with a limited number of training samples below 200, the experimental result reveals an approximately 5\% increase of accuracy for the siamese net based one-shot learning, and a 10\% increase for five-shot learning.
\subsubsection{Others}
There are also a number of other variants for bearing fault diagnostics, either based on novel DL frameworks or mixture of multiple DL methods listed above. For example, in \cite{LAMSTAR}, a new large memory storage retrieval (LAMSTAR) neural network is proposed with 1 input layer, 40 input SOM modules as hidden layers, and 1 decision SOM module as the output layer. More accurate classification results compared to the conventional CNN are reported at various operating conditions, especially at low speeds. In \cite{DN Mix}, the DBN and SAE are applied simultaneously to identify the presence of a bearing fault. Other examples include a mixture of CNN and DBN \cite{CNN DBN Mix}, a deep residual network (DRN) \cite{DRN1, DRN2}, a deep stack network \cite{DSN1}, a RNN based auto-encoders\cite{RNN5}, sparse filtering \cite{Sparse_Filtering}, etc.
\begin{table*}[]
\centering
    \caption{Comparison of Classification Accuracy on Case Western Reserve University Bearing Dataset with Different DL Algorithms.}
    \resizebox{\linewidth}{!}{
\begin{tabular}{ccccccc}
\toprule
Reference                     & \begin{tabular}[c]{@{}c@{}}Feature extraction \\ algorithms\end{tabular}      & \begin{tabular}[c]{@{}c@{}}No. hidden \\ layers\end{tabular} & Classifier                                                                       & Characteristics                                                                & \begin{tabular}[c]{@{}c@{}}Training sample \\ percentage\end{tabular} & Average accuracy \\ 
\midrule
\cite{CNN2}  & Adaptive CNN                                                          & 3                                                            & Softmax                                                                          & Predict fault size                                                             & 50\%                                                                  & 97.90\%           \\
\cite{CNN4}  & CNN                                                                           & 4                                                            & Softmax                                                                          & Noise-resilient                                                                & 90\%                                                                  & 92.60\%           \\
\cite{CNN5}  & CNN                                                                           & 4                                                            & Softmax                                                                          & Sensor fusion                                                                  & 70\%                                                                  & 99.40\%          \\
\cite{CNN6}  & CNN based on LeNet-5                                                          & 8                                                            & FC layer                                                                         & Better feature extraction                                                      & 83\%                                                                & 99.79\%          \\
\cite{CNN7}  & Deep fully connected CNN                                                         & 8                                                            & \begin{tabular}[c]{@{}c@{}} Connectionist \\ temporal classification\end{tabular} & \begin{tabular}[c]{@{}c@{}}Validation with \\ Actual filed test data\end{tabular} & 78\%                                                                & 99.22\%          \\
\cite{CNN8} &  Multi-scale deep CNN      & 9                                                            & Softmax                                                                          & Reduce training time                                                           & 90\%                                                                  & 98.57\%          \\
\cite{CNN9} & \begin{tabular}[c]{@{}c@{}}CNN with\\ training interface\end{tabular}         & 13                                                           & Softmax                                                                          & Adapt to load change                                                           & 96\%                                                                & 95.50\%           \\
\cite{CNN10} & IDS-CNN                                                          & 3                                                            & Softmax                                                                        & Adapt to load change                                                          & 80\%                                                                  & 98.92\%          \\
\cite{CNN11} & CNN-based LiftingNet                                                          & 6                                                            & FC layer                                                                         & Adapt to speed change                                                          & 50\%                                                                  & 99.63\%          \\
\cite{CNN12} & PSPP-CNN                                                          & 9                                                                   & Softmax                                                                          & Adapt to speed change                                                           & 67\%                                                                  & 99.19\%          \\
\cite{CNN13} & AOCNN with SF                                                          & 4                                                                   & Softmax                                                                          & Reduce training set \%                                                          & 5\%                                                                  & 99.19\%          \\
\cite{SA0}   & SAE                                                                           & 3                                                            & ELM                                                                              & Adapt to load change                                                           & 50\%                                                                  & 99.61\%          \\
\cite{SA1}   & SAE                                                                           & 3                                                            & ELM                                                                              & Reduce training time                                                           & 50\%                                                                  & 99.83\%          \\
\cite{SA2}   & Stacked denoising AE                                                    & 3                                                            & N/A                                                                              & Noise-resilient                                                                & 50\%                                                                  & 91.79\%          \\
\cite{SA3}   & SDAE                                                                        & 3                                                            & Softmax                                                                          & Noise-resilient                                                                & 80\%                                                                  & 99.83\%          \\
\cite{SA7}   & Ensemble deep AE                                                       & 3                                                            & Softmax                                                                          & Better feature extraction                                                      & 67\%                                                                & 99.15\%          \\
\cite{SA8}   & Deep wavelet AE                                                         & 3                                                            & ELM                                                                              & Reduce training time                                                           & 67\%                                                                & 95.20\%           \\
\cite{SA9}   & Stack sparse AE                                                                           & 2                                                & N/A                                                                              & Data compression                                                               & N/A                                                                   & 97.47\%          \\
\cite{SA10}   & SAE-local connection network    & 2                                                            & Softmax                                                                          & Shift-invariant features                                                       & 25\%                                                                  & 99.92\%          \\
\cite{SA11}  & SAE                                                                           & 3                                                            & SVM                                                                              & Online diagnosis                                                               & N/A                                                                   & 95.10\%           \\
\cite{SA12}  & SDAE                                                                          & 8                                                            & Gath-Geva (GG)                                                                   & Noise-resilient                                                                & N/A                                                                   & 93.30\%           \\
\cite{SA13}  & Winner-take-all AE                                                                          & 2                                            & Gath-Geva (GG)                                                                   & Noise-resilient                                                                & N/A                                                                   & 97.27\%           \\
\cite{DBN3}  & \begin{tabular}[c]{@{}c@{}}dual-tree\\ complex wavelet\end{tabular}           & 5                                                            & N/A                                                                              & Adaptive DBN                                                                   & 67\%                                                                & 94.38\%          \\
\cite{DBN4}  & DBN                                                                           & 2                                                            & Softmax                                                                          & Adapt to load change                                                           & N/A                                                                   & 98.80\%           \\
\cite{DBN5}  & \begin{tabular}[c]{@{}c@{}}DBN with\\ ensemble learning\end{tabular}          & 4                                                            & Sigmoid                                                                          & Accurate \& robust                                                             & N/A                                                                   & 96.95\%       \\  
\cite{RNN3}  & CNN-LSTM   & 3    & Softmax  & Accurate  & 83\%     & 99.60\%   
\\
\cite{RNN4} & Deep RNN   & 3    & N/A  & Accurate  & 60\%     & 94.75\% 
\\
\cite{GAN4} & DCGAN     & 8    & SVM  & Data augmentation  & 96\%     & 86.33\%
\\
\cite{GAN1} & CatAAE    & 11    & Softmax  & Adapt to load changes  & 91\%     & 90.68\%
\\
\cite{GAN2} & A2CNN     & 27    & Softmax  & Domain adaptation  & N/A\%     & 99.21\%
\\
\cite{GAN3} & GAN+SDAE  & 8    & Softmax \emph{et al.}  & Data augmentation  & 50-78\%     & 99.20\%
\\
\bottomrule
\end{tabular}}
\label{tab:DL_comp}
\end{table*}

\section{Discussions on Deep Learning Algorithms for Bearing Fault Diagnosis}
\subsection{Automated Feature Extraction and Selection}
As opposed to feature engineering of ML algorithms, which manually selects features that preserve the discriminative characteristics of the data, the DL based algorithms can learn the discriminative feature representation directly from input data in an \emph{end-to-end} manner. The DL based approach does not require human expertise or prior knowledge of the problem, and is therefore advantageous in bearing fault diagnosis, where it is sometimes challenging to determine the fault characteristic features accurately. Specifically, DL methods perform feature learning from raw data and classification in a simultaneous and intertwined manner, as illustrated in the cluster visualization results of multiple convolutional layers in Fig. \ref{fig_visual}. A glimpse of the clustering effect can be observed in convolutional layer C2; and it becomes increasingly apparent in later convolutional layers. For comparison reasons, many DL based papers also present results using classical ML methods with human engineered features for bearing fault detection. The majority of DL based methods are reported to outperform traditional ML methods, especially in the presence of external noise and frequent change of operating conditions.
%
%


%
%
\subsection{Comparison of Different DL Algorithms for Bearing Fault Diagnostics}
Thus far, several types DNN architectures and their applications to bearing fault diagnostics have been extensively discussed, and TABLE \ref{tab:DL_sum1} briefly describes the pros and cons of the commonly used deep learning approaches in the field of bearing fault diagnostics. The decision to choose which specific DL algorithm or which specific variant can be customized based on the specific setup environment, the data size, and the number and type of sensors installed. Details on algorithm customization and recommendation will be provided in Section VI.
\subsection{Comparison of DL Algorithm Performance using the CWRU dataset}
A systematic comparison of the classification accuracy of different DL algorithms employing the CWRU bearing dataset is presented in TABLE \ref{tab:DL_comp}. As can be readily observed, the minimum number of hidden layers for all of the networks is 2, indicating the complete network has at least 4 layers incorporating the input and output layers. The maximum hidden layer size can be as large as 13 in \cite{CNN11}, representing a very deep network that requires more time in the training process. The selection criterion for the number of hidden layers is to count the layers that are part of the model's architecture, while excluding the input and the output layer. Based on this criterion, in a CNN we count each convolutional layer and each pooling layer as an effective hidden layer, and disregard any of the dropout layer, since it is a regularization technique that only affects the training process (during evaluation, it is not active, otherwise the weights of the network will be larger than normal). For a GAN, we count all of the hidden layers in both the generator and the discriminator.

The test accuracy of all of the DL algorithms are above 95\%, which validates the feasibility and effectiveness of applying deep learning to bearing fault diagnostics. 
However, it is worthwhile to mention that these specific values of test accuracy cannot be used as the sole indicator to compare the effectiveness of different algorithms for the following reasons:
\begin{enumerate}
    \item \emph{Generalization:} Some of the DL methods with an astonishing accuracy over 99\% are generally applied on a very specific dataset at a fixed operating condition, i.e., when the motor speed is 1,797 rpm and the load is 2 hp. However, this accuracy may suffer significantly under the influence of noise and variation of the motor's speed and load, which unfortunately can be a common issue in practical applications. This is in spite of the relatively strong robustness to noise disturbances of the original DL algorithms (CNN, SAE, DBN, etc.), and their capabilities to learn fault features through a general-purpose learning architecture. It is also reported in \cite{CNN4} that the conventional CNN has a better built-in denoising mechanism compared to other classical DL algorithms such as AE. Due to this limitation, some papers applied the stacked denoising AE (SDAE) \cite{SA2, SA3, SA12} to increase AE's noise resilience under a small SNR, i.e., SNR = 5 or 10.
    \item \emph{Unbalanced Sampling:} Regarding the selection of training samples from the CWRU dataset, many papers did not guarantee a balanced sampling, which means the ratio of data samples selected from the healthy condition and the faulty condition is not close to 1:1. In case of a significant unbalance, accuracy should not be used as the only metric to evaluate an algorithm \cite{GAN2}. Compared with accuracy, other metrics, such as \emph{precision}, \emph{recall} and \emph{F1-Score}, should be introduced to provide more details for evaluating the reliability of a fault identification network. In addition, if the majority of training set are data from the healthy condition, many of the learnt features cannot fully indicate various fault conditions. Therefore, provided that the training data is highly unbalanced, it would be very challenging to apply the DL classifier trained with laboratory data to identify a bearing fault in practical applications, even if the DL framework adopts transfer learning with domain adaptations. 
    \item \emph{Randomness:} Even when these DL methods are using the same dataset to perform classification, the percentage of training data and test data can be different, which unavoidably affects the trustability of the comparison between different approaches. What's more, even if this data distribution is identical, the training and test data might be randomly selected from the CWRU bearing dataset. Therefore this comparison is not performed on the common ground, since the classification accuracy is subject to change even with the same algorithm due to the randomness in selecting the training and the test set. 
    \item \emph{Accuracy saturation:} Most of the existing DL algorithms can achieve an excellent classification accuracy of over 95\% using the CWRU dataset, even with the classical CNN without any add-on architectures, which indicates that this dataset contains relatively simple features that can be easily extracted by a variety of DL methods. In fact, all of the bearing defects in the CWRU dataset are manually drilled or engraved, which are much easier to detect than the realistic bearing spalls or general roughness due to aging. Therefore, various perturbations adding on the original dataset needs to be performed to evaluate more advanced functionalities of DL algorithms, i.e., the CWRU data combined with random noise to test an algorithm's denoising capability. 
\end{enumerate}

All of the factors above would make the classification accuracy of different DL algorithms less convincing.
\section{Suggestions, Challenges, and Future Work Directions}

\subsection{Recommendations and Suggestions}
The successful implementation of machine learning and deep learning algorithms on bearing fault diagnostics can be attributed to the strong correlations among features that follow the law of physics. For engineers and researchers considering applying ML or DL methods to solve their bearing detection problems at hand, the authors suggest the following sequences to make the best algorithm selection.
\begin{enumerate}
    \item \emph{Setup environment:} The first thing we recommend is to thoroughly examine the working environment and all of the possible operating conditions, for example, indoor or outdoor, operating at a fixed operating point or multiple speeds and loads. For the simplest case with an indoor and a single operating point setup, some classical ML methods or even the frequency based analytical model should suffice. For applications that are more prone to external disturbances or having multiple operating points, such as motors fed by VFD converters in electric vehicles, more advanced deep learning approaches should be employed. Specifically, when the workbench is exposed to a noisy environment, which induces a relatively small SNR, certain denoising blocks and extra hidden layers should be added to increase the noise-resiliency and robustness of the deep neural net.
    \item \emph{Sensors:} Then we would need to check the number and type of sensors to be mounted close to the bearing. For the traditional frequency based and classical ML methods, one or two vibrations sensors mounted close to the bearing should be sufficient. For deep learning based approaches, due to the fact that many algorithms such as CNN are mainly developed for computer vision to handle 2-D image data, multiple 1-D time-series data obtained by multiple sensors in the bearing setup need to be stacked together to form this 2-D data. Alternatively, some prepossessing functions, such as the wavelet packet decomposition (WPD), need to be applied before the data is transferred to the deep neural net. Therefore, it would be better to have more than two vibration sensors installed at the same time. In addition, other types of sensors such as acoustic emission and stator current can be installed to form a multi-physics dataset to further improve the accuracy and robustness of the proposed classifier, especially in the midst of frequent and abrupt shifts of operating conditions.
    \item \emph{Data size:} If the size of the collected dataset is not sufficient to train a deep learning algorithm with a good level of generalization, specific algorithms should be selected that can make the most out of the data and computing resources available. For example, dataset augmentation techniques such as GAN, and data random sampling techniques with replacement such as Boostrapping, can be readily implemented. Before actually collecting data from the bearing setup, it is advised to interpret the required sample size beforehand \cite{SampleSizeSel} by considering how accurate the classification result needs to be.
    With a small labeled dataset, another promising routine is to leverage the unlabeled dataset, if possible, and apply the semi-supervised learning paradigm by combining the supervised and unsupervised learning approaches.
 \end{enumerate}
 \subsection{Current Challenges}
Despite the extensive effort and the large number of academic papers devoted to this field, there are still some major challenges that need to be tackled to successfully apply ML and DL algorithms to real-world applications:
 \begin{enumerate}
    \item \emph{Knowledge transfer from laboratories to the real world:} The majority of work included in this review is using publicly available dataset collected from laboratories setups to train their customized ML or DL algorithms. However, it would be ideal to be able to transfer the learned network structure and parameters to detect bearing faults from previously unseen setups, and a very promising example would be learning to predict naturally occuring bearing faults in the real-world by only using data collected from artificial faults in the lab. However, there are still many technique details that need to be tackled to accomplish this ambitious goal.
    \item \emph{Limited labels:} For bearing fault detection, it is often times much easier to collect a large amount of data than to accurately obtain their corresponding labels, and this is especially the case for those faults that evolved naturally over time. Specifically, it is not easy to determine precisely when the first trace of a fault shows up and how long it lasts at the incipient stage.
    \item \emph{Data imbalance:} In certain occasions it can be challenging or expensive to collect a sufficient amount of data at various bearing faulty conditions to effectively train DL algorithms, while the vast majority of data collected would be at the healthy condition, which in fact does not significantly contribute to training an effective and robust bearing fault classifier.
    \item \emph{Noisy data:} Most of the existing work employing DL techniques for bearing fault diagnosis relies on vibration data collected from accelerometers in a laboratory environment. However, in real industrial scenarios such as wind turbines, a large amount of environmental vibration, resonance, or noise may take place. Therefore, it is still an open question if these DL algorithms, while being trained using the vibration data alone, can still deliver satisfactory fault detection performances if the collected data is contaminated by noise.
 \end{enumerate}
\subsection{Future Work Directions}
Regarding future research directions, the authors suggest the following methodologies and algorithms that might be helpful to address the aforementioned challenges:
 \begin{enumerate}
%
    %
    \item \emph{Transfer learning: Transfer learning is a promising technique to transfer the knowledge and experience learned from existing datasets to help identify unforeseen bearing fault conditions at different setups in real-world applications. Typical transfer learning techniques include domain randomization and domain adaptation, which can effectively increase the diversity of the source domain (existing datasets), and help facilitate faster learning and better performance in the target domain (real-world cases).}
    \item \emph{Semi-supervised learning:} To alleviate the problem of ``limited labels'', semi-supervised learning can be utilized to make full use of the limited labeled data and the massive unlabeled data. One potential routine is to employ the variational encoder based deep generative model perform variational inference on data with limited labels.
    \item \emph{Data augmentation:} Data augmentation techniques such as GAN can be introduced solve the ``data imbalance and scarcity'' issue by generating more ``fake'' faulty data to facilitate the training process of DL algorithms. Despite this promising feature, it has been reported in \cite{GAN3} that the accuracy actually declined for some classifiers after incorporating the generated data into the training process. The authors in \cite{GAN3} thus concludes that ``the quality of generated spectrum samples'' generated by GAN ``isn't good enough to provide auxiliary information''. Therefore, it would be interesting to explore more powerful generative models, such as BigGAN, to address this open issue.
    \item \emph{Few-Shot learning:} Another way to address the ``data imbalance and scarcity'' problem is to adapt few-shot learning algorithms to achieve a reasonable classification accuracy using a substantially smaller amount of data. This can be combined with transfer learning and domain adaptation to facilitate the use of deep learning for bearing fault diagnostics at the industry level.
    \item \emph{Explainability:} Rigorous interpretations of DL in general is not well developed as compared with  classical ML methods. Several references, such as \cite{CNN_TL2}, and \cite{CNN_TL1}, attempted to visualize the learnt CNN kernel to interpret its physical meanings. 
    These studies have provided intuitions on the explainability of DL, but more in-depth investigations and their adaptability to bearing fault diagnostics are necessary.
    \item \emph{Sensor fusion:} To solve the potential problem of``noisy data'', it might be worthwhile to deploy other types of sensors, such as the load cell, the current sensor, and the acoustic emission sensor, etc., and apply sensor fusion techniques to synthesize these data and improve the robustness of bearing fault diagnosis. Specifically, the use of acoustic sensors should be advocates since it is reported in \cite{LAMSTAR} that in comparison with vibration signals, acoustic emission signals ``have certain advantages in detecting incipient faults, capturing and representing''. Some existing work on applying acoustic signals to train ML and DL algorithms can be found in \cite{SVM10} and \cite{LAMSTAR}, respectively.
 \end{enumerate}
\section{Conclusions}
In this paper, a systematic review is presented on the existing literature employing deep learning algorithms to bearing fault diagnostics. Special emphasis is placed on deep learning based approaches that has spurred the interest of the research community over the past five years. It is demonstrated that, despite the fact that deep learning algorithms require a large dataset to train, they can automatically perform adaptive feature extractions on the bearing data without any prior expertise on fault characteristic frequencies or operating conditions, making them promising candidates to perform real-time bearing fault diagnostics. A comparative study is also conducted comparing the performance of many DL algorithm variants using the common CWRU bearing dataset. Finally, detailed recommendations and  suggestions are provided in regards to choosing the most appropriate type of DL algorithm for specific application scenarios. Future research directions are also discussed to better facilitate the transition of DL algorithms from laboratory tests to real-world applications.
%





%

\balance

\end{document}